%% file: main.tex
\documentclass[conference]{IEEEtran}
\IEEEoverridecommandlockouts
\pdfoutput=1

\usepackage{float}
\usepackage[table,xcdraw]{xcolor}
\usepackage{hyperref}
\hypersetup{
    colorlinks=true,
    linkcolor=black,
    filecolor=black,
    citecolor=black,
    urlcolor=black,
    pdftitle={Overleaf Example},
    pdfpagemode=FullScreen,
    }
\usepackage{cite}
\usepackage{amsmath,amssymb,amsfonts}
\usepackage{algorithmic}
\usepackage{graphicx}
\usepackage{textcomp}
\usepackage{xcolor}
\usepackage{makecell}
\usepackage{booktabs}
\usepackage{tabularx}
\usepackage{cmap}
\usepackage{fancyhdr}
\usepackage[T1]{fontenc}
\usepackage{multirow}
\usepackage{pifont}
\usepackage{pbox}
\usepackage{amsmath}

\DeclareMathOperator*{\argmin}{min}
\def\BibTeX{{\rm B\kern-.05em{\sc i\kern-.025em b}\kern-.08em
    T\kern-.1667em\lower.7ex\hbox{E}\kern-.125emX}}

\usepackage{todonotes} 

\newcommand{\bl}{\textcolor[rgb]{0, 0, 0}}

\fancypagestyle{copyright}{
\lfoot{978-1-6654-8045-1/22/\$31.00 \copyright2022 IEEE} 
}

\begin{document}
\title{AER: Auto-Encoder with Regression for Time Series Anomaly Detection}

\author{\IEEEauthorblockN{Lawrence Wong}
\IEEEauthorblockA{\textit{MIT}\\
Cambridge, USA \\
lcwong@mit.edu}
\and
\IEEEauthorblockN{Dongyu Liu}
\IEEEauthorblockA{\textit{MIT}\\
Cambridge, USA \\
dongyu@mit.edu}
\and
\IEEEauthorblockN{Laure Berti-Equille}
\IEEEauthorblockA{
\textit{IRD ESPACE-DEV}\\
Montpellier, France \\
laure.berti@ird.fr}
\and
\IEEEauthorblockN{Sarah Alnegheimish}
\IEEEauthorblockA{\textit{MIT}\\
Cambridge, USA \\
smish@mit.edu}
\and
\IEEEauthorblockN{Kalyan Veeramachaneni}
\IEEEauthorblockA{
\textit{MIT}\\
Cambridge, USA \\
kalyanv@mit.edu}
}

\maketitle

\thispagestyle{copyright}
\input{0_abstract}


\input{1_introduction}
\input{2_related_work}
\input{3_definition}
\input{4_methods}
\input{5_experimental_results}
\input{6_conclusion}

\bibliographystyle{abbrv}
\bibliography{main}


\end{document}

%% file: 0_abstract.tex
\begin{abstract}

Anomaly detection on time series data is increasingly common across various industrial domains that monitor metrics in order to prevent potential accidents and economic losses. However, a scarcity of labeled data and ambiguous definitions of anomalies can complicate these efforts. Recent unsupervised machine learning methods have made remarkable progress in tackling this problem using either single-timestamp predictions or time series reconstructions. While traditionally considered separately, these methods are not mutually exclusive and can offer complementary perspectives on anomaly detection. This paper first highlights the successes and limitations of prediction-based and reconstruction-based methods with visualized time series signals and anomaly scores. We then propose AER (Auto-encoder with Regression), a joint model that combines a vanilla auto-encoder and an LSTM regressor to incorporate the successes and address the limitations of each method. Our model can produce bi-directional predictions while simultaneously reconstructing the original time series by optimizing a joint objective function. Furthermore, we propose several ways of combining the prediction and reconstruction errors through a series of ablation studies. Finally, we compare the performance of the AER architecture against two prediction-based methods and three reconstruction-based methods on 12 well-known univariate time series datasets from NASA, Yahoo, Numenta, and UCR. The results show that AER has the highest averaged F1 score across all datasets (a \bl{23.5\%} improvement compared to ARIMA) while retaining a runtime similar to its vanilla auto-encoder and regressor components. Our model is available in Orion\footnote{The AER model is available in Orion: \url{https://github.com/sintel-dev/Orion}
}, an open-source benchmarking tool for time series anomaly detection.

\end{abstract}

%% file: 1_introduction.tex
\section{Introduction}

\begin{figure}[t]
    \includegraphics[width=\linewidth]{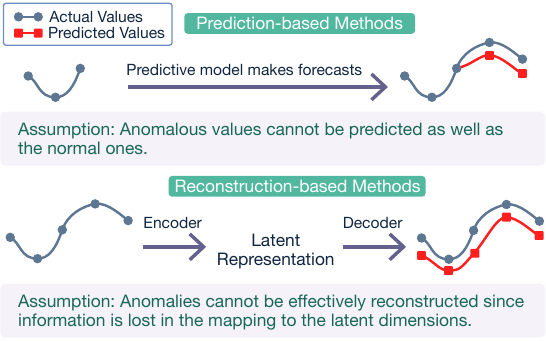}
\vspace{-0.5cm}
\caption{A comparison between prediction-based and reconstruction-based methods for anomaly detection in 
\bl{time series} data. Prediction-based methods learn a predictive model fitted to the given time series data. Reconstruction-based methods learn a model to capture the given time series data's latent structure and then reconstruct the data \bl{from} their latent representations.}
\label{1_methods}
\vspace{-0.7cm}
\end{figure}

Time series data is consistently generated and collected across various industries -- examples include stock prices in finance, vital signs in healthcare, and retail sales in business. 
Effective monitoring and use of time series data are essential for increasing efficiency and productivity. In addition, analysis of \bl{time series} data can extrapolate recurring patterns to predict future occurrences. Anomaly detection, an important task within time series analysis, explicitly aims to identify unexpected events. This research is increasingly relevant due to its broad applications in detecting crucial issues, such as financial fraud in trading networks~\cite{financial_fraud}, medical problems in electrocardiograms~\cite{medical_issues,cheng2021vbridge}, and ecosystem disturbances in satellite signals~\cite{disturbance}.

While the criteria differ across domains, anomalies in time series typically exhibit one of three identifiable patterns: point, contextual, or collective \cite{anomaly_definition}. Point anomalies are singular data points that suddenly deviate from the normal range of the series. For example, a sensor malfunction is one common cause of point anomalies. Collective anomalies are a series of consecutive data points that are considered anomalous as a whole. Finally, contextual anomalies are groups of data points that fall within the series' normal range but do not follow expected temporal patterns. 

Time series also exhibit unique properties that complicate anomaly detection. First, the temporality of time series implicates a correlation or dependence between each consecutive observation \cite{kusuma_doherty_ramchandran}. Second, the dimensionality of each observation influences the computational cost, imposing limitations on the modeling method. For example, modeling methods for multivariate datasets with more than one channel face the curse of dimensionality since they need to capture correlations between observations on top of temporal dependencies \cite{chalapathy_chawla}. Third, noise due to minor 
\bl{sensor fluctuations} during the process of capturing the signal can impact the performance \cite{tuzlukov_cheng}. The pre-processing stages must minimize noise to prevent models from confusing this noise with anomalies. Finally, time series are often non-stationary. They have statistical properties that change over time, like seasonality, concept drift, and change points, that can easily be mistaken for anomalies.

Existing machine learning methods for anomaly detection on time series can be either prediction-based or reconstruction-based (Fig.~\ref{1_methods}). Prediction-based methods train a model to learn previous patterns in order to forecast future observations \cite{geiger2020tadgan}. An observation is anomalous when the predicted value deviates significantly from the actual value. Prediction-based methods are good at revealing point anomalies but tend to produce more false detection \cite{prediction-based-false-positives}. On the other hand, reconstruction-based methods learn a latent low-dimensional representation to reconstruct the original input \cite{geiger2020tadgan}. This method assumes that anomalies are rare events that are lost in the mapping to the latent space. Hence, regions that cannot be effectively reconstructed are considered anomalous. In our experiments, we observed that reconstruction-based methods tend to be more effective than prediction-based methods at identifying contextual and collective anomalies.

This paper proposes a new architecture -- an auto-encoder with regression (AER) model -- that leverages the successes and addresses the limitations faced by each method type. This architecture trains a reconstruction-based auto-encoder with a prediction-based regression component using a joint objective function. As a result, the model can produce both reconstruction-based and prediction-based anomaly scores (likelihood of an abnormal observation). This paper also explores several ways to calculate and combine scores 
\bl{to} address several limitations of existing methods. Briefly, the contributions of this paper are as follows:
\begin{itemize}
    \item We identified several successes and limitations of prediction-based and reconstruction-based methods using visualized examples. 
    
    \item We propose a novel architecture -- auto-encoder with regression (AER) -- that leverages the successes of prediction-based and reconstruction-based methods for anomaly detection on time series data.
    
    \item We introduce the idea of masking anomaly scores created from the smoothing function to reduce start-of-sequence false-positive predictions. We applied masking to every baseline method and compared the method's performance to that of its unmasked counterpart.
    
    \item We present bi-directional anomaly scores, which combine prediction-based anomaly scores in the forward and reverse directions. This method addresses the limitation of missing forecasts faced by prediction-based methods.
    
    \item We demonstrated that AER outperformed five other baseline methods in anomaly detection on 12 time series datasets\footnote{Scripts to reproduce the results are available in  \url{https://github.com/sintel-dev/aer-paper}}. In addition, ablation studies show that the AER model achieved a 
    \bl{23.5\%} improvement in averaged F1 score compared to the baseline ARIMA model while retaining a runtime similar to its vanilla auto-encoder and LSTM regressor components.

\end{itemize}

The structure of the paper is as follows: Section II provides an overview of the existing pipeline and approaches for time series anomaly detection. Section III formally defines the problem, and Section IV documents the successes and limitations of existing methods. Section V introduces our solution, including the AER framework, smoothing function masking, and bi-directional \bl{scoring}. Finally, Sections VI and VII evaluate the proposed framework, discuss the results, and summarize the key findings.

%% file: 2_related_work.tex
\section{Related Work}

\subsection{Anomaly Detection Pipelines}
Anomaly detection aims to find a set of anomalous intervals from either univariate or multivariate time series data. It is usually an unsupervised task \bl{due to the lack of labeled data}. Recent work by Sintel \cite{Alnegheimish_2022} formalized this task as an end-to-end pipeline consisting of pre-processing, modeling, and post-processing stages. The pre-processing stage first transforms the raw data into suitable inputs for the models. The modeling stage then predicts or reconstructs the input to get the expected output. Finally, the post-processing stage finds discrepancies between the expected and real inputs. The methodology for finding these discrepancies significantly impacts the anomalies identified by this stage. Hence, our work focuses on the limitations in the post-processing stage for prediction-based and reconstruction-based methods. Understanding these limitations also enables us to make appropriate changes to the \mbox{modeling stage}.

\subsection{Machine Learning-Based Approaches}

\textbf{Prediction-based approaches} generally use the deviation between the predicted and actual values to identify anomalies. Autoregressive Integrated Moving Average (ARIMA) \cite{arima} and Long Short Term Memory Recurrent Neural Network with Non-parametric Dynamic Thresholding (LSTM-DT) \cite{lstm_ndt} are well-known examples of prediction-based approaches. ARIMA uses lags and lagged forecast errors to predict future values. Statistical models like ARIMA require the user to have extensive domain knowledge about the time series data in order to adjust the parameters appropriately. Machine learning-based methods like LSTM-DT tend to require less domain knowledge. In the modeling stage, the method uses a separate LSTM neural network to model each channel in order to facilitate granular system control and mitigate errors from high-dimensionality outputs. In the post-processing stage, the method combines \bl{an} exponentially-weighted average function with a non-parametric dynamic thresholding technique to detect anomalous intervals. Our work examines the limitations of the post-processing stage in LSTM-DT pipeline.

\textbf{Reconstruction-based approaches} learn a latent low-dimensional representation to reconstruct the original input. These methods assume that the latent space prioritizes capturing common patterns within the dataset. Rare events like anomalies are not captured in the latent representation and are less likely to be accurately reconstructed. Principal Component Analysis (PCA) \cite{pca}, LSTM Auto-Encoders (LSTM-AE) \cite{lstm_ae}, and LSTM Variational Auto-Encoders (LSTM-VAE) \cite{lstm_vae} are examples of reconstruction-based approaches. PCA is a dimensionality-reduction technique limited to linear reconstructions and fails to leverage spatial-temporal correlation in multivariate settings. LSTM-AE is an auto-encoder built from LSTM layers that learns a latent space representation for the input. \bl{The size of the latent space needs to be calibrated to capture generalizable patterns while avoiding noise and anomalies.} LSTM-VAE introduces regularization in the latent space using a probabilistic encoder and decoder. However, these methods tend to overfit to the training data, which results in decreased performances \cite{geiger2020tadgan}.

Generative Adversarial Network (GAN) is another reconstruction-based approach to address the overfitting issue. 
\bl{This form of adversarial learning offers regularization to the reconstruction errors}. An early example is MAD-GAN \cite{li2019madgan}, which uses spatial-temporal correlation and other dependencies among multiple variables to capture non-linear latent interactions. TadGAN \cite{geiger2020tadgan} is another GAN-based approach trained with cycle consistency loss to address model instability issues and allow for better reconstruction of time series data. It also proposes several methods in the post-processing stage to calculate reconstruction-based anomaly scores. Similar to prediction-based methods, our work examines post-processing steps presented by TadGAN for reconstruction-based approaches.

Zhao et al. propose MTAD-GAT, a multivariate anomaly detection model that optimizes a joint loss of forecasting-- and reconstruction--based models~\cite{zhao2020multivariate}. The architecture of MTAD-GAT differs from AER (our work) where MTAD-GAT is a graph attention network in comparison to AER that includes a bidirectional LSTM network. Moreover, Zhao et al. apply additional preprocessing steps to clean the data. Specifically, they apply Spectral Residual (SR) anomaly detection method~\cite{ren2019time} to filter out anomalous regions. In this work, we limit preprocessing to data scaling, imputing, and detrending. Furthermore, our approach still operates in an unsupervised setting where there is no prior knowledge about the anomalies in the dataset and no hyperparameter tuning, preventing information leakage.
Lastly, we provide analysis to understand why the combination of prediction-based and reconstruction-based anomaly scores can be beneficial in predicting point and collective anomalies.





%% file: 3_definition.tex
\section{ML-Based Anomaly Detection Pipeline}

\begin{figure}[t]
\centering
\includegraphics[width=\linewidth]{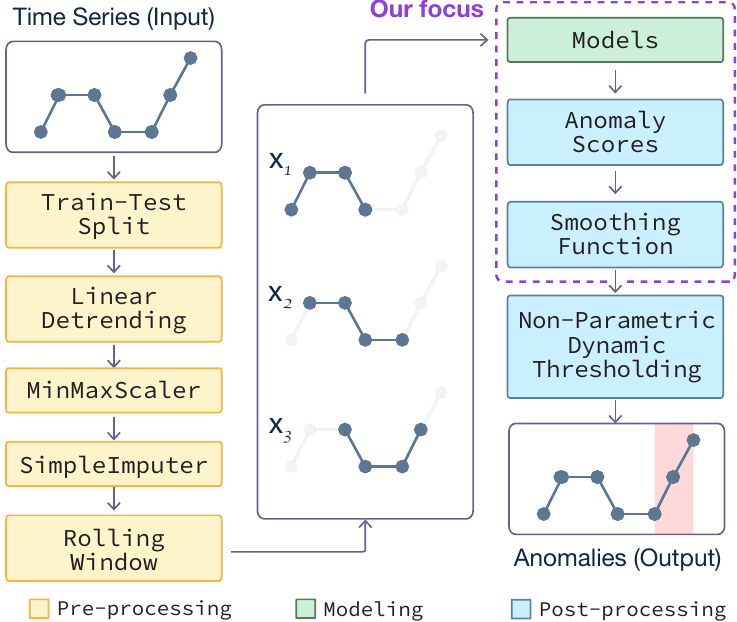}
\caption{The pipeline for anomaly detection on time series data consists of pre-processing, modeling, and post-processing stages. Our work focuses on models, anomaly scores, and the smoothing function \bl{steps} of the pipeline.}
\label{2:pipeline}
\vspace{-0.5cm}
\end{figure}

Unsupervised time series anomaly detection aims to find a set of anomalous intervals given time series with one or more channels. Ideally, each interval captures an unexpected behavior that deviates from the expected patterns in the signal. 
\bl{This section first formulates the anomaly detection task into a sequence of steps (Fig.~\ref{2:pipeline}) similar to Alnegheimish et al.'s work~\cite{Alnegheimish_2022} and then critically analyzes existing methods to learn their strengths and weaknesses.}

\subsection{Pre-processing Stage}
The time series signal is pre-processed into inputs suitable for models similar to Geiger et al.'s work \cite{geiger2020tadgan}. The time series $t$ with $d$ number of channels \bl{is} divided into train and test splits. The train split is used to learn the parameters for subsequent transformations. Both splits are detrended, as necessary, by fitting and subtracting a least-square fit. Then, \bl{the} values of each split are min-max normalized to the range \texttt{[-1, 1]}. Finally, any missing values are imputed with the mean. Let $T$ be the total number of observations in the split without loss of generality. A rolling window with window size $n$ and step size $1$ creates $T-n$ number of inputs $\mathbf{x}_i = \{t_i, t_{i+1}, ..., t_{i+n-1}\}$ such that $i$ represents the index of the first observation in the window. It is worth noting that pre-processing varies based on application scenarios, and the above summary only covers the most common steps.

\subsection{Modeling Stage}
The input and output depend on the type of anomaly detection model. Each input $\mathbf{x}_i \in \mathbb{R}^{n \times d}$ has $n$ observations based on the window size (default to $n=100$ for reconstruction-based models and $n=250$ for prediction-based models) with $d$ channels. In the case of multivariate inputs, separate models are trained for each channel to ensure traceability \cite{lstm_ndt}. Usually, one channel is selected as the model's target channel. For example, many-to-one prediction-based models will produce single timestep predictions $f_i \in \mathbb{R}$ for index $i$ of the target channel. On the other hand, many-to-one reconstruction-based models reconstruct the entire target channel and produce a sequence $y_{i: i+n-1} \in \mathbb{R}^{n}$ with the same starting index $i$ as the input $\mathbf{x}_i$.

\begin{table*}[t]
\centering
\resizebox{\linewidth}{!}{
\renewcommand{\arraystretch}{2}
\begin{tabular}{|c|
>{\columncolor[HTML]{F6A6A9}}c l|
>{\columncolor[HTML]{E2EFD9}}c l|}
\hline
\cellcolor[HTML]{E7E6E6}\small{\textbf{Anomaly Scores}} &
  \multicolumn{2}{c|}{\cellcolor[HTML]{F6A6A9}\small{\textbf{Limitations (L)}}} &
  \multicolumn{2}{c|}{\cellcolor[HTML]{E2EFD9}\small{\textbf{Successes (S)}}} \\ \hline
 &
  \multicolumn{1}{c|}{\cellcolor[HTML]{F6A6A9}\textbf{PL1}} &
  \makecell[lc]{High anomaly scores at the early indices often result in false-\\positive predictions.} &
  \multicolumn{1}{c|}{\cellcolor[HTML]{E2EFD9}} &
  \\ \cline{2-3}
 &
  \multicolumn{1}{c|}{\cellcolor[HTML]{F6A6A9}\textbf{PL2}} &
  \makecell[lc]{Low prediction-based anomaly scores for contextual anomalies \\ with simple patterns result in false-negative predictions.} &
  \multicolumn{1}{c|}{\cellcolor[HTML]{E2EFD9}} &
  \\ \cline{2-3}
\multirow{-3}{*}{Prediction-based (P)} &
  \multicolumn{1}{c|}{\cellcolor[HTML]{F6A6A9}\textbf{PL3}} &
  \makecell[lc]{Missing prediction-based anomaly scores at the early indices \\result in false-negative predictions.} &
  \multicolumn{1}{c|}{\multirow{-3}{*}{\cellcolor[HTML]{E2EFD9}\textbf{PS1}}} &
  \multirow{-3}{*}{\makecell[lc]{Prediction-based anomaly scores are better at capturing \\point anomalies than reconstruction-based anomaly scores.}} \\ \hline
 &
  \multicolumn{1}{c|}{\cellcolor[HTML]{F6A6A9}} &
  &
  \multicolumn{1}{l|}{\cellcolor[HTML]{E2EFD9}\textbf{RS1}} &
  \makecell[lc]{Reconstruction-based anomaly scores are better at capturing \\contextual and collective anomalies.} \\ \cline{4-5} 
\multirow{-2}{*}{Reconstruction-based (R)} &
  \multicolumn{1}{c|}{\multirow{-2}{*}{\cellcolor[HTML]{F6A6A9}\textbf{RL1}}} &
  \multirow{-2}{*}{\makecell[lc]{Reconstruction-based anomaly scores reducing peaks for point \\anomalies result in false-negative predictions.}} &
  \multicolumn{1}{l|}{\cellcolor[HTML]{E2EFD9}\textbf{RS2}} &
  \makecell[lc]{Reconstruction-based DTW anomaly scores are better at \\capturing anomalies than AD and PD anomaly scores.} \\ \hline
\end{tabular}
}
\vspace{.20cm}
\caption{Overview of successes (S) and limitations (L) for prediction-based (P) and reconstruction-based (R) methods.}
\label{3:overview_sl}
\vspace{-0.7cm}
\end{table*}

\subsection{Post-processing Stage -- Computing Anomaly Scores}

\bl{The computation of anomaly scores differs between prediction-based and reconstruction-based models since they produce different outputs.}

\textbf{Prediction-based models} produce a one-step forecast in the forward direction $f_{i+n}$ at index $i+n$ given the input $\mathbf{x}_i$ starting at index $i$. Only forecasts \bl{$f_{i}$} for indices \bl{$i \in [n+1, T]$} can be computed since prediction-based models require at least $n$ observations 
\bl{to} forecast the first value at the index $n+1$. The absolute error between the sequence of forecasts in the forward direction $f$ and the time series $t$ creates the prediction-based anomaly score $\alpha_{p}$ as defined in Eq. (\ref{eq:1}).
\begin{equation}
    \alpha_p(t, f) = 
     \begin{cases}
       0 & i \in [1, n+1) \\
       |t_i - f_i| & i \in \bl{[n+1, T]}
     \end{cases}
    \label{eq:1}
\end{equation}

\textbf{Reconstruction-based models} reconstruct a sequence of values $y_{i:i+n-1}$ of one channel given the input $\mathbf{x}_i$ starting at index $i$. Each index $i$ in the time series signal has multiple reconstructed values since that index occurs in multiple sequences of $y_{i:i+n-1}$. 
\bl{The median of the collection of reconstructed values is used as the final value for index $i$ since using the median achieves better performance than using the mean~\cite{geiger2020tadgan}.} Unlike prediction-based anomaly scores, reconstruction-based anomaly scores can be calculated for every index. The reconstruction-based anomaly scores can be calculated in three ways given sequences $t$ and $y$: point-wise differencing, area differencing, or dynamic time warping.

\textit{ Point-wise differencing (PD)}. The reconstruction-based PD anomaly score $\alpha_{r, p}$ defined in Eq. (\ref{eq:2}) takes the absolute error between the time series $t$ and the reconstructed value $y$ at every index $i$.
\begin{equation}
    \alpha_{r, p}(t, y) = |t_i - y_i| \hspace{1cm} i \in \bl{[1, T]}
    \label{eq:2}
\end{equation}

\textit{Area differencing (AD)}. The reconstruction-based AD anomaly score $\alpha_{r, a}$ defined in Eq. (\ref{eq:3}) is created using a fixed length window size that measures the similarity between local regions.
\begin{equation}
    \alpha_{r, a}(t, y) = \frac{1}{2l}\left|\int_{i-l}^{i+l} t_i - y_i\hspace{1mm}dt\right| \hspace{.5cm} i \in \bl{[1, T]}
    \label{eq:3}
\end{equation}
The similarity is measured as the average difference between areas beneath two curves of length $2l$ calculated using the trapezoidal rule ($l=10$ by default).

\textit{Dynamic Time Warping (DTW)}. The reconstruction-based DTW anomaly score $\alpha_{r, d}$ defined in Eq. (\ref{eq:4}) created with dynamic time warping allows for many-to-many mapping between two sequences that are locally out of phase \cite{dtw}. DTW creates a cost matrix $C \in \mathbb{R}^{2l\times2l}$ such that each $(i, j)$ coordinate represents the distance $c_q$ between $t_i$ and $y_j$. 
\begin{equation}
    \alpha_{r, d}(t, y) = \argmin_{C}\left[\frac{1}{Q}\sqrt{\sum_{q=1}^{Q}c_q}\right]
    \label{eq:4}
\end{equation}
Dynamic programming solves for the optimal warp path $C^*$ with the minimum warp distances between $t$ and $y$.

\bl{Exponentially weighted moving average (EWMA)~\cite{EWMA} with a smoothing window of $0.1T$ is applied to both prediction-based and reconstruction-based anomaly scores to \mbox{reduce noise.}}

\subsection{Post-processing Stage -- Identifying Anomalous Sequences}

\bl{Hundman et al. \cite{lstm_ndt} used the locally adaptive thresholding function to identify anomalous intervals from the anomaly scores}. This \bl{function} uses a sliding window to compute local thresholds, merges continuous observations to create anomalous sequences, and mitigates false-positives by pruning anomalies.

Let $\alpha$ be the sequence of anomaly scores with a maximum size of length $T$ (one score for each observation). The window size defaults to $\frac{T}{3}$ with a step size of $\frac{T}{3 * 10}$ to optimally identify anomalies. The adaptive threshold for each sliding window is four standard deviations from the window's mean. Observations with scores that exceed that threshold are identified as anomalous. Consecutive anomalous time steps are joined together to create $K$ anomalous sequences. 
\bl{Hundman et al.~\cite{lstm_ndt} additionally employed a pruning method to reduce the number of false positives.} Let $K_{max}^{(i)}$ represent the maximum anomaly score in each anomalous sequence $K^{(i)}$. The maxima are sorted in descending order, and the decrease percentage change $p^{(i)}$ is calculated between $K_{max}^{(i)}$ and $K_{max}^{(i+1)}$. At the sequence $K^{(j)}$ whose percentage change $p^{(j)}$ does not exceed an empirically defined threshold $\theta$ (defaults to 0.13), that sequence and all subsequent sequences are reclassified as normal, i.e., all sequences \mbox{between \texttt{[j, K]}}.

\section{Critical Analysis of Existing Methods}

\begin{figure}
\includegraphics[width=\linewidth]{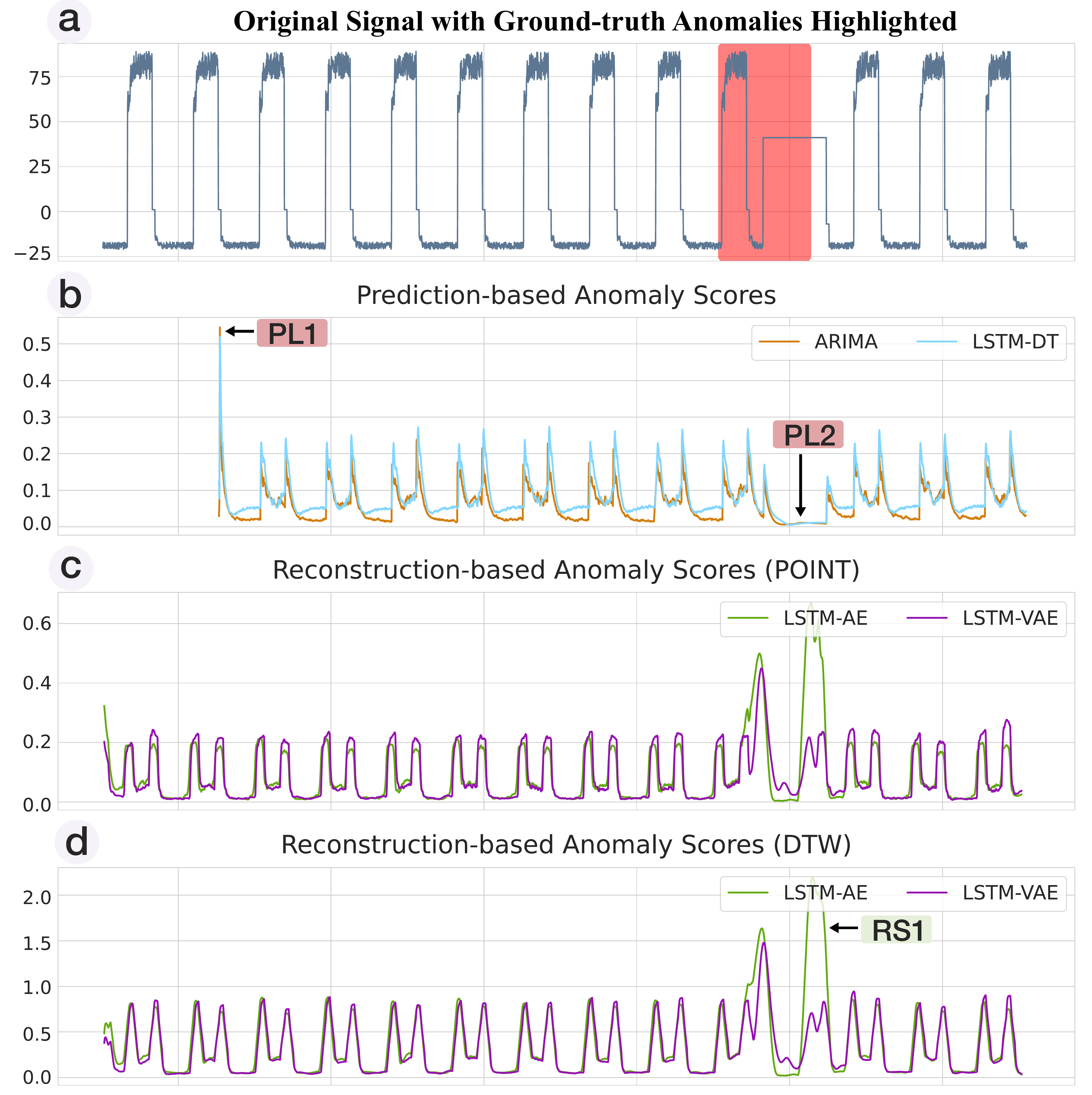}
\caption{Anomaly scores for the \texttt{art\_daily\_flatmiddle} signal from the \texttt{artificialWithAnomaly} dataset with one contextual anomaly. Prediction-based anomaly scores are high near the beginning (PL1) and low for contextual anomalies with simple patterns (PL2). On the other hand, all variations of reconstruction-based anomaly scores could capture the simple contextual anomaly (RS1).}
\label{3:interval}
\vspace{-0.5cm}
\end{figure}

\begin{figure}
\includegraphics[width=\linewidth]{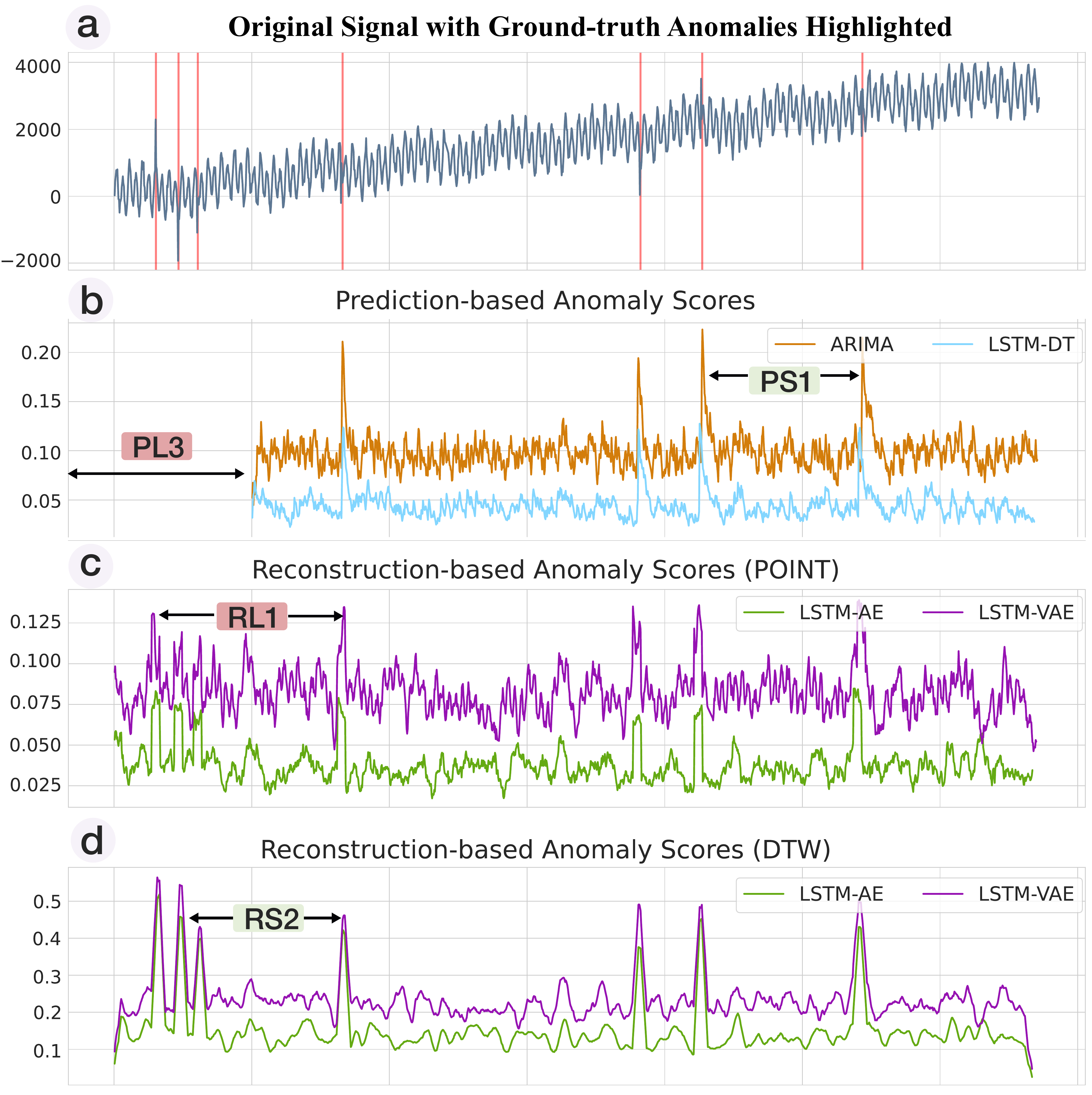}
\caption{Anomaly scores for the \texttt{A3Benchmark-TS11} signal from the \texttt{YAHOOA3} dataset with multiple point anomalies. Prediction-based anomaly scores are better at identifying point anomalies than reconstruction-based anomaly scores (PS1) but fail to find anomalies at the start of the signal (PL3). Of the variations in calculating reconstruction-based anomaly scores, DTW was the best at capturing the point anomalies (RS2).}
\label{3:spikes}
\vspace{-0.5cm}
\end{figure}

Despite some minor differences, most prediction-based (P) and reconstruction-based (R) methods follow the same course presented in Fig.~\ref{2:pipeline}. Both method types generally have their successes (S) and limitations (L), which are summarized in Table~\ref{3:overview_sl}. 
Fig. \ref{3:interval} and \ref{3:spikes} illustrate the time series signal and the anomaly scores produced by each method \bl{in} real-world datasets to \bl{understand each of their behavior better}. In each figure, graph (a) shows the time series signal (blue) with the ground truth anomalous intervals (red). Graph (b) shows the prediction-based anomaly score $\alpha_{p}$ produced by ARIMA (orange) and LSTM-DT (\bl{sky blue}) models. Graphs (c,d) correspond to the reconstruction-based PD $\alpha_{r, p}$ and DTW $\alpha_{r, d}$ anomaly scores for LSTM-AE (green) and LSTM-VAE (purple) models.

\textbf{PL1}: High anomaly scores at the early indices often result in false-positive predictions. This error is likely the byproduct of using the exponential weighted moving average function to smooth the anomaly scores. The function requires at least the same number of observations as the size of the smoothing window before it can produce stable anomaly scores. While this limitation occurs in many signals, an example is seen in the prediction-based anomaly scores from the \texttt{art\_daily\_flatmiddle} signal (see PL1 in Fig. \ref{3:interval}(b)).

\textbf{PL2}: Low prediction-based anomaly scores for contextual anomalies with simple patterns result in false-negative predictions. The cyclic pattern in prediction-based anomaly scores suggests that the models could not fully capture the structure, especially at the change point in the time series. However, in this case, the contextual anomaly is a simple pattern. Therefore, the models can easily forecast the pattern, resulting in nearly zero anomaly scores at the interval. Hence, the adaptive threshold failed to find the contextual anomaly (see PL2 in Fig. \ref{3:interval}(b)).

\textbf{PL3}: Missing prediction-based anomaly scores at the early indices result in false-negative predictions (see PL3 in Fig.~\ref{3:spikes}(b)). This limitation occurs only in prediction-based models since they require at least $n$ observations to forecast the first value at index $n+1$. This behavior usually results in false-negative predictions for signals with anomalies occurring at the beginning, mainly from datasets like \texttt{YAHOOA3} with a decent number of point anomalies at the start of the time series.


\textbf{PS1}: Prediction-based anomaly scores are better at capturing point anomalies than reconstruction-based anomaly scores. For example, prediction-based anomaly scores showed more prominent peaks at anomalies than reconstruction-based anomaly scores for the \texttt{A3Benchmark-TS11} signal from the \texttt{YAHOOA3} dataset. As a result, the locally adaptive thresholding function can quickly identify anomalies using prediction-based anomalies, resulting in higher F1 scores for datasets like \texttt{YAHOOA3} with more point anomalies (see PS1 in Fig.~\ref{3:spikes}(b)).

\textbf{RL1}: Reconstruction-based anomaly scores reducing peaks for point anomalies result in false-negative predictions. The reconstruction-based anomaly scores are calculated from the median of all predicted values for index $i$. Since some reconstructed outputs are better at capturing the point anomalies than others, the median value is closer to the true value at index $i$. This calculation lowers the anomaly scores such that the window-based threshold no longer captures those point anomalies, since the scores are now closer to the window's mean (see RL1 in Fig.~\ref{3:spikes}(c)).

\textbf{RS1}: Reconstruction-based anomaly scores are better at capturing contextual and collective anomalies. For example, reconstruction-based anomaly scores from the \texttt{art\_daily\_flatmiddle} signal spiked while prediction-based anomaly scores remained close to zero at the contextual anomaly (see RS1 in Fig.~\ref{3:interval}(d)). This behavior occurs for prediction-based anomaly scores since the contextual anomaly pattern was easy to model. On the other hand, reconstruction-based models struggled to recreate the entire interval, since the model tries to reconstruct values from simple anomalous intervals and complex non-anomalous intervals. The sudden shift from an intricate cyclic pattern to a simple pattern results in high reconstruction-based anomaly scores.

\textbf{RS2}: Reconstruction-based DTW anomaly scores are better at capturing anomalies than AD and PD anomaly scores. Reconstruction-based anomaly scores for the \texttt{A3Benchmark-TS11} signal show that reconstruction-based DTW anomaly scores are less noisy than reconstruction-based PD anomaly scores (see RS2 in Fig.~\ref{3:spikes}(d)). The success of DTW scores is attributed to the method's ability to handle shifts in the alignment of two series. The ablation study by Geiger et al. \cite{geiger2020tadgan} also reports that DTW slightly outperforms the other two reconstruction error types.


Our observations show that prediction-based and reconstruction-based anomaly scores have successes and limitations that complement one another. For example, we observe from our experiments that prediction-based anomaly scores have an easier time identifying point anomalies but produce relatively more false positives. On the other hand, reconstruction-based anomaly scores have an easier time identifying contextual and collective anomalies but produce relatively more false negatives. Therefore, our method strives to address these limitations and leverage strengths from both types of models as an alternative solution for anomaly detection in time series.  

%% file: 4_methods.tex
\section{AER: Auto-Encoder with Regression}

\begin{table*}[t]
\centering
\resizebox{\linewidth}{!}{
\renewcommand{\arraystretch}{1.2}

\begin{tabular}{|c|l|cc|cccc|ccccc|c|}
\hline
\rowcolor[HTML]{E7E6E6} 
\cellcolor[HTML]{E7E6E6} &
  \multicolumn{1}{c|}{\cellcolor[HTML]{E7E6E6}\textbf{Source}} &
  \multicolumn{2}{c|}{\cellcolor[HTML]{E7E6E6}\textbf{NASA}} &
  \multicolumn{4}{c|}{\cellcolor[HTML]{E7E6E6}\textbf{YAHOO}} &
  \multicolumn{5}{c|}{\cellcolor[HTML]{E7E6E6}\textbf{NAB}} &
  \textbf{UCR} \\ \cline{2-14} 
\rowcolor[HTML]{FFF2CC} 
\multirow{-2}{*}{\cellcolor[HTML]{E7E6E6}\textbf{Datasets}} &
  \multicolumn{1}{c|}{\cellcolor[HTML]{FFF2CC}\textbf{Name}} &
  \multicolumn{1}{c|}{\cellcolor[HTML]{FFF2CC}\textbf{MSL}} &
  \textbf{SMAP} &
  \multicolumn{1}{c|}{\cellcolor[HTML]{FFF2CC}\textbf{A1}} &
  \multicolumn{1}{c|}{\cellcolor[HTML]{FFF2CC}\textbf{A2}} &
  \multicolumn{1}{c|}{\cellcolor[HTML]{FFF2CC}\textbf{A3}} &
  \textbf{A4} &
  \multicolumn{1}{c|}{\cellcolor[HTML]{FFF2CC}\textbf{Art}} &
  \multicolumn{1}{c|}{\cellcolor[HTML]{FFF2CC}\textbf{AdEx}} &
  \multicolumn{1}{c|}{\cellcolor[HTML]{FFF2CC}\textbf{AWS}} &
  \multicolumn{1}{c|}{\cellcolor[HTML]{FFF2CC}\textbf{Traffic}} &
  \textbf{Tweets} &
  \textbf{UCR} \\ \hline
\cellcolor[HTML]{E7E6E6} &
  \cellcolor[HTML]{FFF2CC}\textbf{Synthetic} &
  \multicolumn{1}{c|}{No} &
  No &
  \multicolumn{1}{c|}{No} &
  \multicolumn{1}{c|}{Yes} &
  \multicolumn{1}{c|}{Yes} &
  Yes &
  \multicolumn{1}{c|}{Yes} &
  \multicolumn{1}{c|}{No} &
  \multicolumn{1}{c|}{No} &
  \multicolumn{1}{c|}{No} &
  No &
  Mix \\ \cline{2-14} 
\multirow{-2}{*}{\cellcolor[HTML]{E7E6E6}\textbf{Properties}} &
  \cellcolor[HTML]{FFF2CC}\textbf{\# Signals} &
  \multicolumn{1}{c|}{27} &
  53 &
  \multicolumn{1}{c|}{67} &
  \multicolumn{1}{c|}{100} &
  \multicolumn{1}{c|}{100} &
  100 &
  \multicolumn{1}{c|}{6} &
  \multicolumn{1}{c|}{5} &
  \multicolumn{1}{c|}{17} &
  \multicolumn{1}{c|}{7} &
  10 &
  250 \\ \hline
\cellcolor[HTML]{E7E6E6} &
  \cellcolor[HTML]{FFF2CC}\textbf{Point (\textit{len=1})} &
  \multicolumn{1}{c|}{0} &
  0 &
  \multicolumn{1}{c|}{68} &
  \multicolumn{1}{c|}{33} &
  \multicolumn{1}{c|}{935} &
  833 &
  \multicolumn{1}{c|}{0} &
  \multicolumn{1}{c|}{0} &
  \multicolumn{1}{c|}{0} &
  \multicolumn{1}{c|}{0} &
  0 &
  3 \\ \cline{2-14} 
\multirow{-2}{*}{\cellcolor[HTML]{E7E6E6}\textbf{\# Anomalies}} &
  \cellcolor[HTML]{FFF2CC}\textbf{Collective (\textit{len\textgreater{}1})} &
  \multicolumn{1}{c|}{36} &
  67 &
  \multicolumn{1}{c|}{110} &
  \multicolumn{1}{c|}{167} &
  \multicolumn{1}{c|}{4} &
  2 &
  \multicolumn{1}{c|}{6} &
  \multicolumn{1}{c|}{11} &
  \multicolumn{1}{c|}{30} &
  \multicolumn{1}{c|}{14} &
  33 &
  247 \\ \hline
\cellcolor[HTML]{E7E6E6} &
  \cellcolor[HTML]{FFF2CC}\textbf{Anomalous Points} &
  \multicolumn{1}{c|}{7766} &
  54696 &
  \multicolumn{1}{c|}{1669} &
  \multicolumn{1}{c|}{466} &
  \multicolumn{1}{c|}{943} &
  837 &
  \multicolumn{1}{c|}{2418} &
  \multicolumn{1}{c|}{795} &
  \multicolumn{1}{c|}{6312} &
  \multicolumn{1}{c|}{1560} &
  15651 &
  49363 \\ \cline{2-14} 
\multirow{-2}{*}{\cellcolor[HTML]{E7E6E6}\textbf{\# Data Points}} &
  \cellcolor[HTML]{FFF2CC}\textbf{Total Points} &
  \multicolumn{1}{c|}{132046} &
  562800 &
  \multicolumn{1}{c|}{94866} &
  \multicolumn{1}{c|}{142100} &
  \multicolumn{1}{c|}{168000} &
  168000 &
  \multicolumn{1}{c|}{24192} &
  \multicolumn{1}{c|}{7965} &
  \multicolumn{1}{c|}{67644} &
  \multicolumn{1}{c|}{15662} &
  158511 &
  19353766 \\ \hline
\end{tabular}
}
\vspace{.20cm}
\caption{High-level overview of all 12 benchmark datasets.}
\label{5:datasets}
\vspace{-0.5cm}
\end{table*}

\begin{figure}
\centering
\includegraphics[width=\linewidth]{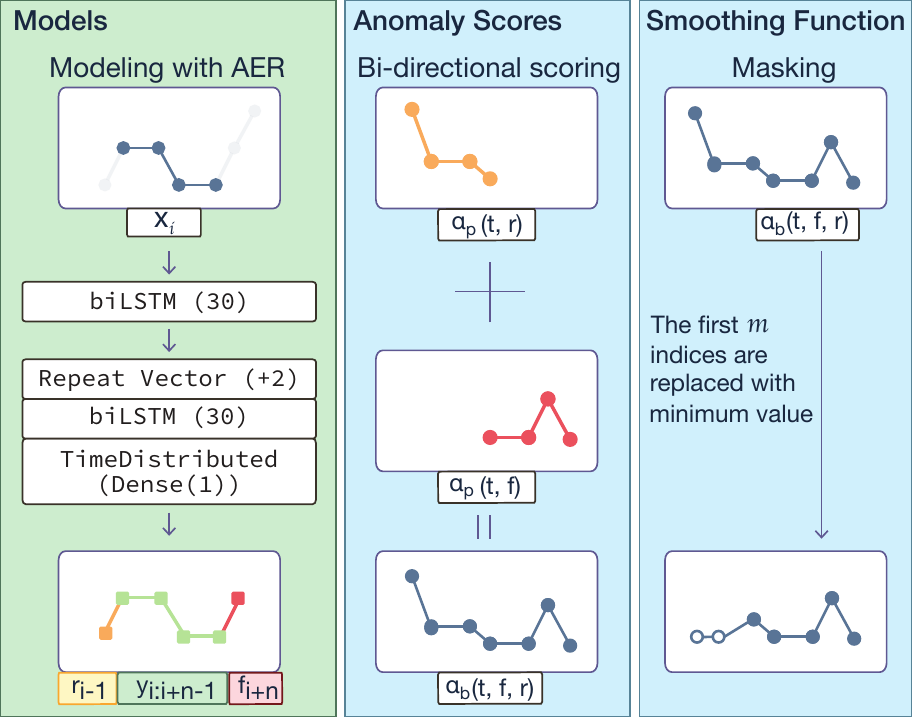}
\caption{Our solution targets three steps in the anomaly detection pipeline: \bl{models, anomaly scores, and smoothing function}. AER is a joint model consisting of an LSTM auto-encoder and regressor capable of producing forecasts and reconstructing the input sequence. Second, bi-directional scoring combines prediction-based anomaly scores in the forward and reverse directions to address the issue of missing anomaly scores (PL3). Finally, masking in the smoothing step replaces the values of the anomaly scores at the first few indices to address start-of-sequence false-positive predictions (PL1).}
\label{4:solution}
\end{figure}

Our solution has three components targeting the \bl{models, anomaly scores, and smoothing function} steps in the anomaly detection pipeline, as summarized in Fig.~\ref{4:solution}.

\subsection{Modeling Stage}

The AER model borrows ideas from LSTM-AE and LSTM-DT to produce prediction-based and reconstruction-based anomaly scores simultaneously. The goal is to combine the strengths of both types of methods while overcoming some of their limitations.

The input to the model is  $\mathbf{x}_i \in \mathbb{R}^{n \times d}$ with $n$ observations and $d$ channels. Like other auto-encoder architectures, AER consists of an encoder and a decoder. While AER uses a regular encoder, the decoder reconstructs $n+2$ instead of $n$ observations by increasing the number of units of the repeated vector layer by two. This minor change allows the model to create an output consisting of three components: the one-step reverse prediction $r_{i-1} \in \mathbb{R}$, the reconstructed sequence $y_{i:i+n-1} \in \mathbb{R}^{n}$, and the one-step-ahead prediction \bl{$f_{i+n} \in \mathbb{R}$}.

The loss function (Eq. \ref{eq:5}) is divided into prediction and reconstruction portions. The prediction loss $V_{pred}$ is the average of the mean squared error between the pairs of true and prediction values in the reverse ($t_{i-1}, r_{i-1}$) and forward direction \bl{($t_{i+n}, f_{i+n}$)}. Likewise, the reconstruction loss $V_{rec}$ is the mean squared error between the time series \bl{$t_{i:i+n-1}$} and the reconstructed sequence $y_{i:i+n-1}$. The contribution of the prediction and reconstruction loss is determined by $\gamma \in [0,1]$. The full objective function is defined as follows:
\bl{\begin{equation}
\begin{split}
    Loss & =
    \frac{\gamma}{2} V_{pred}(t_{i-1}, r_{i-1})
    + \frac{\gamma}{2} V_{pred}(t_{i+n}, f_{i+n}) \\
    &\qquad+ (1-\gamma) V_{rec}(t_{i:i+n-1}, y_{i:i+n-1})
\end{split}
\label{eq:5}
\end{equation}}
By default, the hyperparameters are $n=100$ observations per input and $\gamma=0.5$ to give equal importance to the prediction and reconstruction losses. One biLSTM layer with $b=30$ units is used for both the encoder and decoder. The latent space is the same dimension as the last hidden state of the bidirectional LSTM layer, which is $2b$.

\subsection{Post-processing Stage: Masking}
To overcome the false-positive predictions created from the exponential weighted moving average smoothing function (PL1), we introduce masking. The proposed solution is to mask $m$ indices from the start of the sequence with some value. Our observations show that using the minimum anomaly scores as the masking value produced the best results. By default, $m$ is equal to $0.01T$ (size of smoothing window) where $T$ is the time series length.

\subsection{Post-processing Stage: Bi-Directional Scoring}
Bi-directional anomaly scores target the missing start of sequence anomaly scores since prediction-based methods require at least $n$ observations to make the first forecast (PL3). A solution is to produce anomaly scores using the  sequence of predictions in the forward direction $f$ and in the reverse direction $r$. The anomaly scores created using $r$ can fill in the missing prediction-based anomaly scores produced by $f$. Again, let $\alpha_p$ denote the function to calculate prediction-based anomaly scores. Prediction-based anomaly scores are calculated in the forward direction $\alpha_{p}(t, f)$ for indices \bl{$i \in [n+1, T]$} and in the reverse direction $\alpha_p(t, r)$ for indices \bl{$i \in [1, T-n]$}. If masking is used, then the first $m$ values of $\alpha_{p}(t, f)$ are replaced with zeros, and the first $m$ values of $\alpha_p(t, r)$ are replaced with $min(\alpha_p(t, r))$. Then, the scores  $\alpha_{p}(t, f)$ are padded with $n$ zeros in the beginning while $\alpha_{p}(t, r)$ are padded with $n$ zeros at the end to align the anomaly scores.
\bl{\begin{equation}
    \resizebox{0.91\hsize}{!}{
    $\alpha_b(t, f, r) = 
     \begin{cases}
       \alpha_p(t, r) & i \in [1, n+m+1)\\
       \frac{1}{2}\alpha_p(t, r) + \frac{1}{2}\alpha_p(t, f) & i \in [n+m+1,  T-n+1)\\
       \alpha_p(t, f) & i \in [T-n+1, T]
     \end{cases}$
    }
    \label{eq:6}
\end{equation}}
The bi-directional anomaly scores $\alpha_b$ defined in Eq. (\ref{eq:6}) consist of averages of both scores in overlapping intervals and the max between both scores in non-overlapping intervals.

\subsection{Post-processing Stage: Combination Scores}
The bi-directional prediction-based anomaly scores $\alpha_b$ and reconstruction-based anomaly errors $\alpha_r$ can be used to create the combined anomaly scores $\alpha_{c}$.

\subsubsection{Prediction-based Only (PRED)}
The combined anomaly scores $\alpha_c$ are calculated using only the bi-directional prediction-based anomaly scores. 
\begin{equation}
    \alpha_{c}(t, r, y, f) = \alpha_{b}(t, f, r).
\end{equation}

\subsubsection{Reconstruction-based Only (REC)}
The combined anomaly scores $\alpha_c$ are calculated using only the reconstruction-based anomaly scores. The calculation of reconstruction-based anomaly scores defaults to using DTW since it outperforms reconstruction-based PD and AD (RS2).
\begin{equation}
    \alpha_{c}(t, r, y, f) = \alpha_{r, d}(t, y).
\end{equation}

\subsubsection{Convex (SUM)}
The combined anomaly scores $\alpha_c$ are calculated using a convex combination with parameter weight $\beta$ that controls the two errors' relative importance (by default $\beta=0.5$). Both prediction-based and reconstruction-based anomaly scores are min-max scaled to between \texttt{[0, 1]} before the combination. 
\begin{equation}
    \alpha_{c}(t, r, y, f) = (1-\beta)\alpha_{r, d}(t, y) + \beta \alpha_{b}(t, f, r).
\end{equation}

\subsubsection{Product (MULT)}
The combined anomaly scores $\alpha_c$ are calculated using a point-wise product between the two scores to emphasize both scores' high values. $\beta$ controls the relative importance of the two errors (by default $\beta=1$). Both prediction-based and reconstruction-based anomaly scores are min-max scaled to between \texttt{[1, 2]} before the combination. 
\begin{equation}
    \alpha_{c}(t, r, y, f) = \beta\alpha_{r, d}(t, y) \odot \alpha_{b}(t, f, r).
    \vspace{0.15cm}
\end{equation}

%% file: 5_experimental_results.tex
\section{Experimental Results}

The three main points we seek to validate in our experimental study are as follows:
\begin{itemize}
    \item \textbf{RQ1}: Does the AER framework enable us to discover anomalies more efficiently than we can through other approaches?
    
    \item \textbf{RQ2}: What is the impact of smoothing function masking and bi-directional scoring on anomaly detection?
    
    \item \textbf{RQ3}: Do mixture anomaly scores offer additional information compared to using either a prediction-based or reconstruction-based anomaly score on its own?
\end{itemize}

\begin{table*}[t]
\centering
\resizebox{\linewidth}{!}{
\renewcommand{\arraystretch}{1.2}

\begin{tabular}{|l|cc|cccc|ccccc|c|c|}
\hline
\rowcolor[HTML]{E7E6E6} 
\multicolumn{1}{|c|}{\cellcolor[HTML]{E7E6E6}{\color[HTML]{000000} }} &
  \multicolumn{2}{c|}{\cellcolor[HTML]{E7E6E6}{\color[HTML]{000000} \textbf{NASA}}} &
  \multicolumn{4}{c|}{\cellcolor[HTML]{E7E6E6}{\color[HTML]{000000} \textbf{YAHOO}}} &
  \multicolumn{5}{c|}{\cellcolor[HTML]{E7E6E6}{\color[HTML]{000000} \textbf{NAB}}} &
  {\color[HTML]{000000} \textbf{UCR}} &
  \cellcolor[HTML]{E7E6E6}{\color[HTML]{000000} } \\ \cline{2-13}
\rowcolor[HTML]{FFF2CC} 
\multicolumn{1}{|c|}{\multirow{-2}{*}{\cellcolor[HTML]{E7E6E6}{\color[HTML]{000000} \textbf{Models}}}} &
  \multicolumn{1}{c|}{\cellcolor[HTML]{FFF2CC}{\color[HTML]{000000} \textbf{MSL}}} &
  {\color[HTML]{000000} \textbf{SMAP}} &
  \multicolumn{1}{c|}{\cellcolor[HTML]{FFF2CC}{\color[HTML]{000000} \textbf{A1}}} &
  \multicolumn{1}{c|}{\cellcolor[HTML]{FFF2CC}{\color[HTML]{000000} \textbf{A2}}} &
  \multicolumn{1}{c|}{\cellcolor[HTML]{FFF2CC}{\color[HTML]{000000} \textbf{A3}}} &
  {\color[HTML]{000000} \textbf{A4}} &
  \multicolumn{1}{c|}{\cellcolor[HTML]{FFF2CC}{\color[HTML]{000000} \textbf{Art}}} &
  \multicolumn{1}{c|}{\cellcolor[HTML]{FFF2CC}{\color[HTML]{000000} \textbf{AdEx}}} &
  \multicolumn{1}{c|}{\cellcolor[HTML]{FFF2CC}{\color[HTML]{000000} \textbf{AWS}}} &
  \multicolumn{1}{c|}{\cellcolor[HTML]{FFF2CC}{\color[HTML]{000000} \textbf{Traffic}}} &
  {\color[HTML]{000000} \textbf{Tweets}} &
  {\color[HTML]{000000} \textbf{UCR}} &
  \multirow{-2}{*}{\cellcolor[HTML]{E7E6E6}{\color[HTML]{000000} \textbf{Avg. F1 ($\mu \pm \sigma$)}}} \\ \hline
\cellcolor[HTML]{FFFFFF}{\color[HTML]{000000} ARIMA} &
  \multicolumn{1}{c|}{\cellcolor[HTML]{F8696B}{\color[HTML]{000000} 0.442}} &
  \cellcolor[HTML]{F8696B}{\color[HTML]{000000} 0.333} &
  \multicolumn{1}{c|}{\cellcolor[HTML]{9CD5AC}{\color[HTML]{000000} 0.733}} &
  \multicolumn{1}{c|}{\cellcolor[HTML]{F8696B}{\color[HTML]{000000} 0.807}} &
  \multicolumn{1}{c|}{\cellcolor[HTML]{8ACE9D}{\color[HTML]{000000} 0.818}} &
  \cellcolor[HTML]{71C487}{\color[HTML]{000000} 0.700} &
  \multicolumn{1}{c|}{\cellcolor[HTML]{F8696B}{\color[HTML]{000000} 0.353}} &
  \multicolumn{1}{c|}{\cellcolor[HTML]{F86E70}{\color[HTML]{000000} 0.518}} &
  \multicolumn{1}{c|}{\cellcolor[HTML]{63BE7B}{\color[HTML]{000000} \textbf{0.741}}} &
  \multicolumn{1}{c|}{\cellcolor[HTML]{FAB4B7}{\color[HTML]{000000} 0.500}} &
  \cellcolor[HTML]{FBE5E7}{\color[HTML]{000000} 0.567} &
  \cellcolor[HTML]{F8696B}{\color[HTML]{000000} 0.124} &
  {\color[HTML]{000000} 0.553   ± 0.21} \\ \hline
\cellcolor[HTML]{FFFFFF}{\color[HTML]{000000} LSTM-DT} &
  \multicolumn{1}{c|}{\cellcolor[HTML]{FBF1F4}{\color[HTML]{000000} 0.515}} &
  \cellcolor[HTML]{CEEAD8}{\color[HTML]{000000} 0.707} &
  \multicolumn{1}{c|}{\cellcolor[HTML]{ADDCBB}{\color[HTML]{000000} 0.721}} &
  \multicolumn{1}{c|}{\cellcolor[HTML]{63BE7B}{\color[HTML]{000000} \textbf{0.980}}} &
  \multicolumn{1}{c|}{\cellcolor[HTML]{AFDDBD}{\color[HTML]{000000} 0.744}} &
  \cellcolor[HTML]{96D3A7}{\color[HTML]{000000} 0.638} &
  \multicolumn{1}{c|}{\cellcolor[HTML]{F88A8C}{\color[HTML]{000000} 0.400}} &
  \multicolumn{1}{c|}{\cellcolor[HTML]{F8696B}{\color[HTML]{000000} 0.513}} &
  \multicolumn{1}{c|}{\cellcolor[HTML]{63BE7B}{\color[HTML]{000000} \textbf{0.741}}} &
  \multicolumn{1}{c|}{\cellcolor[HTML]{63BE7B}{\color[HTML]{000000} \textbf{0.667}}} &
  \cellcolor[HTML]{FBF8FB}{\color[HTML]{000000} 0.580} &
  \cellcolor[HTML]{B2DEBF}{\color[HTML]{000000} 0.391} &
  {\color[HTML]{000000} 0.633   ± 0.16} \\ \hline
\cellcolor[HTML]{FFFFFF}{\color[HTML]{000000} LSTM-AE} &
  \multicolumn{1}{c|}{\cellcolor[HTML]{FAD5D8}{\color[HTML]{000000} 0.500}} &
  \cellcolor[HTML]{D2EBDB}{\color[HTML]{000000} 0.705} &
  \multicolumn{1}{c|}{\cellcolor[HTML]{FABEC1}{\color[HTML]{000000} 0.610}} &
  \multicolumn{1}{c|}{\cellcolor[HTML]{EEF7F3}{\color[HTML]{000000} 0.866}} &
  \multicolumn{1}{c|}{\cellcolor[HTML]{F87E80}{\color[HTML]{000000} 0.420}} &
  \cellcolor[HTML]{F87173}{\color[HTML]{000000} 0.253} &
  \multicolumn{1}{c|}{\cellcolor[HTML]{FBF2F5}{\color[HTML]{000000} 0.545}} &
  \multicolumn{1}{c|}{\cellcolor[HTML]{63BE7B}{\color[HTML]{000000} \textbf{0.750}}} &
  \multicolumn{1}{c|}{\cellcolor[HTML]{FBE3E6}{\color[HTML]{000000} 0.692}} &
  \multicolumn{1}{c|}{\cellcolor[HTML]{F8696B}{\color[HTML]{000000} 0.457}} &
  \cellcolor[HTML]{F8696B}{\color[HTML]{000000} 0.483} &
  \cellcolor[HTML]{FBFAFD}{\color[HTML]{000000} 0.314} &
  {\color[HTML]{000000} 0.550   ± 0.17} \\ \hline
\cellcolor[HTML]{FFFFFF}{\color[HTML]{000000} LSTM-VAE} &
  \multicolumn{1}{c|}{\cellcolor[HTML]{EFF7F4}{\color[HTML]{000000} 0.526}} &
  \cellcolor[HTML]{FBF0F3}{\color[HTML]{000000} 0.653} &
  \multicolumn{1}{c|}{\cellcolor[HTML]{F99799}{\color[HTML]{000000} 0.575}} &
  \multicolumn{1}{c|}{\cellcolor[HTML]{F99B9D}{\color[HTML]{000000} 0.823}} &
  \multicolumn{1}{c|}{\cellcolor[HTML]{F88789}{\color[HTML]{000000} 0.432}} &
  \cellcolor[HTML]{F8696B}{\color[HTML]{000000} 0.240} &
  \multicolumn{1}{c|}{\cellcolor[HTML]{63BE7B}{\color[HTML]{000000} \textbf{0.667}}} &
  \multicolumn{1}{c|}{\cellcolor[HTML]{B5DFC2}{\color[HTML]{000000} 0.700}} &
  \multicolumn{1}{c|}{\cellcolor[HTML]{F98F91}{\color[HTML]{000000} 0.643}} &
  \multicolumn{1}{c|}{\cellcolor[HTML]{F99699}{\color[HTML]{000000} 0.483}} &
  \cellcolor[HTML]{63BE7B}{\color[HTML]{000000} \textbf{0.590}} &
  \cellcolor[HTML]{FBFCFE}{\color[HTML]{000000} 0.317} &
  {\color[HTML]{000000} 0.554   ± 0.16} \\ \hline
\cellcolor[HTML]{FFFFFF}{\color[HTML]{000000} TadGAN} &
  \multicolumn{1}{c|}{\cellcolor[HTML]{63BE7B}{\color[HTML]{000000} \textbf{0.584}}} &
  \cellcolor[HTML]{FBE1E4}{\color[HTML]{000000} 0.617} &
  \multicolumn{1}{c|}{\cellcolor[HTML]{F8696B}{\color[HTML]{000000} 0.533}} &
  \multicolumn{1}{c|}{\cellcolor[HTML]{FAD6D9}{\color[HTML]{000000} 0.842}} &
  \multicolumn{1}{c|}{\cellcolor[HTML]{F8696B}{\color[HTML]{000000} 0.391}} &
  \cellcolor[HTML]{F98D90}{\color[HTML]{000000} 0.297} &
  \multicolumn{1}{c|}{\cellcolor[HTML]{EAF5F0}{\color[HTML]{000000} 0.571}} &
  \multicolumn{1}{c|}{\cellcolor[HTML]{DAEFE2}{\color[HTML]{000000} 0.677}} &
  \multicolumn{1}{c|}{\cellcolor[HTML]{BFE4CB}{\color[HTML]{000000} 0.720}} &
  \multicolumn{1}{c|}{\cellcolor[HTML]{CCE9D5}{\color[HTML]{000000} 0.581}} &
  \cellcolor[HTML]{8CCF9F}{\color[HTML]{000000} 0.588} &
  \cellcolor[HTML]{F88688}{\color[HTML]{000000} 0.162} &
  {\color[HTML]{000000} 0.547   ± 0.18} \\ \hline
\cellcolor[HTML]{FFFFFF}{\color[HTML]{000000} AER*} &
  \multicolumn{1}{c|}{\cellcolor[HTML]{CBE8D5}{\color[HTML]{000000} 0.541}} &
  \cellcolor[HTML]{63BE7B}{\color[HTML]{000000} \textbf{0.772}} &
  \multicolumn{1}{c|}{\cellcolor[HTML]{63BE7B}{\color[HTML]{000000} \textbf{0.772}}} &
  \multicolumn{1}{c|}{\cellcolor[HTML]{7DC992}{\color[HTML]{000000} 0.959}} &
  \multicolumn{1}{c|}{\cellcolor[HTML]{63BE7B}{\color[HTML]{000000} \textbf{0.896}}} &
  \cellcolor[HTML]{63BE7B}{\color[HTML]{000000} \textbf{0.722}} &
  \multicolumn{1}{c|}{\cellcolor[HTML]{ACDCBA}{\color[HTML]{000000} 0.615}} &
  \multicolumn{1}{c|}{\cellcolor[HTML]{FBE6E9}{\color[HTML]{000000} 0.635}} &
  \multicolumn{1}{c|}{\cellcolor[HTML]{F8696B}{\color[HTML]{000000} 0.621}} &
  \multicolumn{1}{c|}{\cellcolor[HTML]{ADDCBB}{\color[HTML]{000000} 0.606}} &
  \cellcolor[HTML]{CAE8D4}{\color[HTML]{000000} 0.585} &
  \cellcolor[HTML]{63BE7B}{\color[HTML]{000000} \textbf{0.470}} &
  {\color[HTML]{000000} \textbf{0.683   ± 0.14}} \\ \hline
\end{tabular}
}
\vspace{.20cm}
\caption{F1 scores for AER compared to prediction-based and reconstruction-based baseline models. \bl{The highest scores are highlighted in dark green, while the lowest scores are highlighted in dark red per dataset.}}
\label{5:all_scores}
\vspace{-.75cm}
\end{table*}

\subsection{Data Sources}

We use 12 datasets (742 signals) spanning various domains to evaluate the models' generalizability and adaptability. The National Aeronautics and Space Administration (NASA) provided two spacecraft telemetry datasets\footnote{NASA data: \url{https://github.com/khundman/telemanom/}}: Soil Moisture Active Passage (SMAP) and Mars Science Laboratory (MSL) acquired from a satellite and a rover, respectively \cite{lstm_ndt}. Each numeric measurement in the target channel is accompanied by one-hot encoded information about commands sent or received by specific spacecraft modules in a given time window. The Yahoo Webscope Program provided the S5 datasets\footnote{Yahoo data: \url{https://webscope.sandbox.yahoo.com/catalog.php?datatype=s&did=70}} consisting of one set of real production traffic to Yahoo properties (A1) and three synthetic datasets (A2, A3, A4) with varying trends, noise, and pre-specified or random seasonality. The A2 and A3 datasets only contain outliers inserted at random positions, while A4 has outliers and change points. The Numenta Anomaly Benchmark (NAB) provided several datasets\footnote{NAB data: \url{https://github.com/numenta/NAB}} from various domains: artificialWithAnomaly (Art), realAdExchange (AdEx), realAWSCloudwatch (AWS), realTraffic (Traffic), realTweets (Tweets). 
The UCR Time Series Anomaly Archive\footnote{UCR data: \url{https://www.cs.ucr.edu/~eamonn/time_series_data_2018/UCR_TimeSeriesAnomalyDatasets2021.zip}} is a dataset created to address flaws like triviality, unrealistic anomaly density, mislabeled ground truth, and run-to-failure bias faced by popular datasets \cite{Wu_2021}.

Similar to Geiger et al. \cite{geiger2020tadgan}, Table \ref{5:datasets} summarizes basic information about each dataset. It differentiates between real and synthetic datasets and provides the number of signals and anomalies \bl{for each dataset}. Each anomaly is classified as either point or collective, depending on the length of the \bl{anomaly}. Lastly, the total number of anomalous and overall data points are provided for each dataset. 

\begin{figure}[t]
\centering
\includegraphics[width=\linewidth]{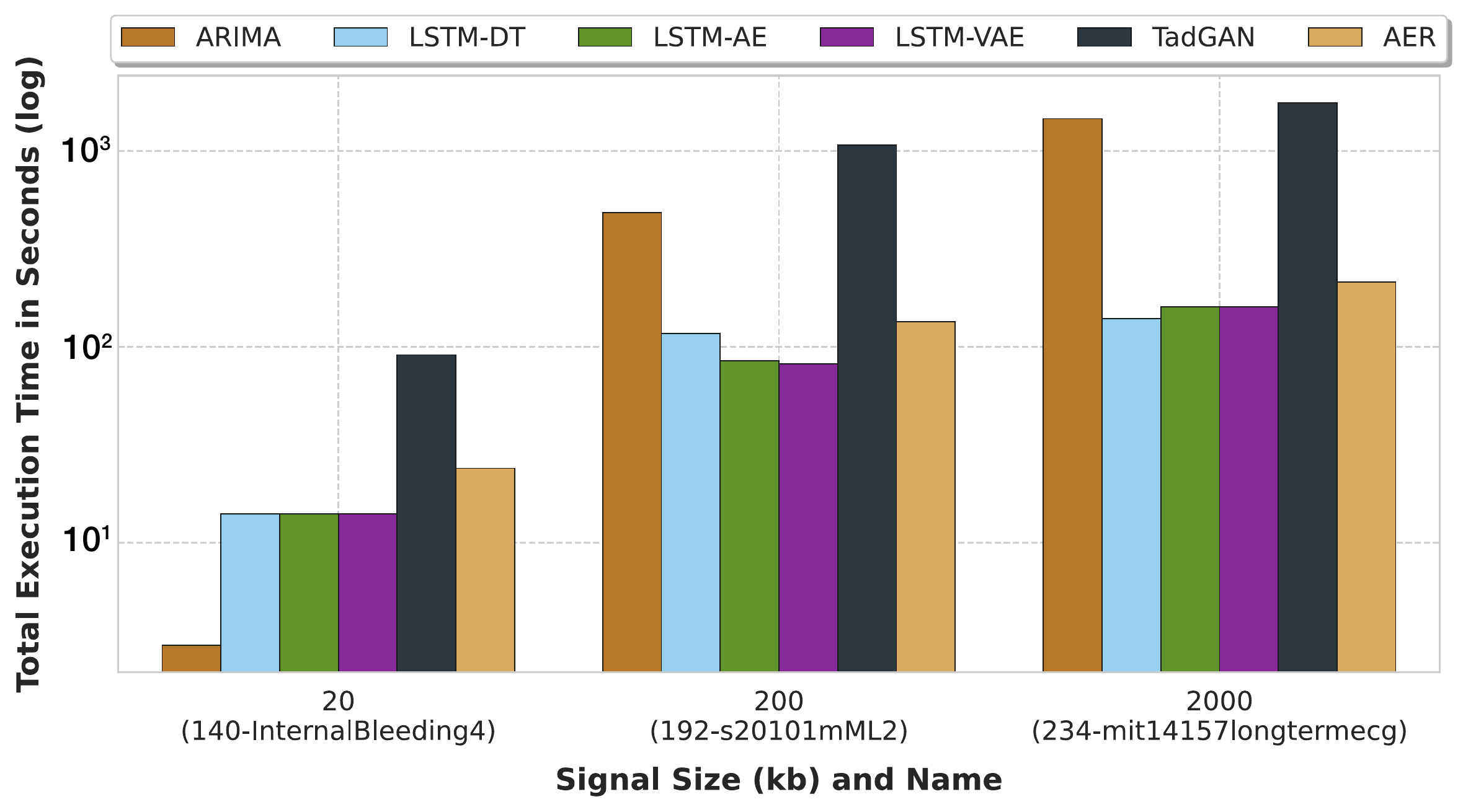}
\caption{
\bl{A comparison between the total execution time in seconds on signals from the UCR dataset with increasing sizes.  The total execution time consists of the time to train the model pipeline (training time) and the time to convert an input into an output (pipeline latency).
}}
\label{5:runtime}
\vspace{-0.3cm}
\end{figure}

\subsection{Evaluation Metrics}
Like Hundman et al. \cite{lstm_ndt} and Geiger et al. \cite{geiger2020tadgan}, the metric used in this study is unweighted contextual F1 scores for each dataset. The motivation is that anomalies are rare and window-based in many real-world application scenarios. The end user's goal is to detect timely true alarms without receiving many false positives. Hence, this evaluation metric is preferable since it prioritizes finding any part of the anomalies. Anomaly scoring is based on overlapping segments: a true positive (TP) if a known anomalous window overlaps any detected windows, a false negative (FN) if a known anomalous window does not overlap any detected windows, and a false positive (FP) if a detected window does not overlap any known \mbox{anomalous region}.

We performed all experiments in an instance of MIT Supercloud \cite{mitsupercloud} with an Intel Xeon Gold 6249 processor, 10 CPU cores, 9 GB RAM per core, and 1 Nvidia Volta V100 GPU. The environment is created using the \textit{anaconda/2022a} module, which includes TensorFlow 2.0. All models are implemented as primitives and benchmarked using Orion \cite{geiger2020tadgan}.

\subsection{Baseline Models}
We compare our solution against the following five state-of-the-art methods:

\textbf{ARIMA} \textit{(Prediction-based)}: Autoregressive Integrated Moving Average \cite{arima} is implemented with the \texttt{StatsModels} library. The hyperparameters are empirically set to \texttt{p=1, d=0, q=0}.

\textbf{LSTM-DT} \textit{(Prediction-based)}: LSTM non-parametric Dynamic Threshold \cite{lstm_ndt} uses two LSTM layers with 80 units and a dropout rate of 0.3. The training hyperparameters were: 35 epochs, batch size of 64, and Adam optimizer.

\textbf{LSTM-AE} \textit{(Reconstruction-based)}: LSTM auto-encoders \cite{lstm_ae} use one LSTM layer with 60 units for the encoder and generator. A time-distributed layer with a dense one-unit layer is used to create the output. 

\textbf{LSTM-VAE} (Reconstruction-based): LSTM variational auto-encoders \cite{lstm_vae} consist of an encoder and a decoder. The encoder uses one shared LSTM layer with 60 units and separate dense layers, each with 60 units, to create the mean and standard deviation vector. The decoder uses a repeat vector layer, an LSTM layer with 60 units, and a time-distributed layer with a dense one-unit layer.

\textbf{TadGAN} \textit{(Reconstruction-based)}: TadGAN \cite{geiger2020tadgan} consists of an encoder and generator that use bi-directional LSTM layers, and critics that use 1D convolution layers. The reconstruction-based anomaly scores can be used in combination with the critic scores to create the final anomaly scores. Geiger et al. \cite{geiger2020tadgan} reported an ablation study merging these scores using summation, product, critic-only, and reconstruction-only combinations. 

\begin{table*}[!ht]
\centering
\resizebox{\linewidth}{!}{
\renewcommand{\arraystretch}{1.2}

\begin{tabular}{lccccccccccccc}
\multicolumn{14}{c}{{\color[HTML]{000000} \textbf{A: Masking and Bi-Directional   Scoring Comparison}}} \\ \hline
\rowcolor[HTML]{E7E6E6} 
\multicolumn{1}{|c|}{\cellcolor[HTML]{E7E6E6}{\color[HTML]{000000} }} &
  \multicolumn{2}{c|}{\cellcolor[HTML]{E7E6E6}{\color[HTML]{000000} \textbf{NASA}}} &
  \multicolumn{4}{c|}{\cellcolor[HTML]{E7E6E6}{\color[HTML]{000000} \textbf{YAHOO}}} &
  \multicolumn{5}{c|}{\cellcolor[HTML]{E7E6E6}{\color[HTML]{000000} \textbf{NAB}}} &
  \multicolumn{1}{c|}{\cellcolor[HTML]{E7E6E6}{\color[HTML]{000000} \textbf{UCR}}} &
  \multicolumn{1}{c|}{\cellcolor[HTML]{E7E6E6}{\color[HTML]{000000} }} \\ \cline{2-13}
\rowcolor[HTML]{FFF2CC} 
\multicolumn{1}{|c|}{\multirow{-2}{*}{\cellcolor[HTML]{E7E6E6}{\color[HTML]{000000} \textbf{Models}}}} &
  \multicolumn{1}{c|}{\cellcolor[HTML]{FFF2CC}{\color[HTML]{000000} \textbf{MSL}}} &
  \multicolumn{1}{c|}{\cellcolor[HTML]{FFF2CC}{\color[HTML]{000000} \textbf{SMAP}}} &
  \multicolumn{1}{c|}{\cellcolor[HTML]{FFF2CC}{\color[HTML]{000000} \textbf{A1}}} &
  \multicolumn{1}{c|}{\cellcolor[HTML]{FFF2CC}{\color[HTML]{000000} \textbf{A2}}} &
  \multicolumn{1}{c|}{\cellcolor[HTML]{FFF2CC}{\color[HTML]{000000} \textbf{A3}}} &
  \multicolumn{1}{c|}{\cellcolor[HTML]{FFF2CC}{\color[HTML]{000000} \textbf{A4}}} &
  \multicolumn{1}{c|}{\cellcolor[HTML]{FFF2CC}{\color[HTML]{000000} \textbf{Art}}} &
  \multicolumn{1}{c|}{\cellcolor[HTML]{FFF2CC}{\color[HTML]{000000} \textbf{AdEx}}} &
  \multicolumn{1}{c|}{\cellcolor[HTML]{FFF2CC}{\color[HTML]{000000} \textbf{AWS}}} &
  \multicolumn{1}{c|}{\cellcolor[HTML]{FFF2CC}{\color[HTML]{000000} \textbf{Traffic}}} &
  \multicolumn{1}{c|}{\cellcolor[HTML]{FFF2CC}{\color[HTML]{000000} \textbf{Tweets}}} &
  \multicolumn{1}{c|}{\cellcolor[HTML]{FFF2CC}{\color[HTML]{000000} \textbf{UCR}}} &
  \multicolumn{1}{c|}{\multirow{-2}{*}{\cellcolor[HTML]{E7E6E6}{\color[HTML]{000000} \textbf{Avg. F1 ($\mu \pm \sigma$)}}}} \\ \hline
\rowcolor[HTML]{FFFFFF} 
\multicolumn{1}{|l|}{\cellcolor[HTML]{FFFFFF}{\color[HTML]{000000} ARIMA}} &
  \multicolumn{1}{c|}{\cellcolor[HTML]{FFFFFF}{\color[HTML]{000000} 0.442}} &
  \multicolumn{1}{c|}{\cellcolor[HTML]{FFFFFF}{\color[HTML]{000000} 0.333}} &
  \multicolumn{1}{c|}{\cellcolor[HTML]{FFFFFF}{\color[HTML]{000000} 0.733}} &
  \multicolumn{1}{c|}{\cellcolor[HTML]{FFFFFF}{\color[HTML]{000000} 0.807}} &
  \multicolumn{1}{c|}{\cellcolor[HTML]{FFFFFF}{\color[HTML]{000000} \textbf{0.818}}} &
  \multicolumn{1}{c|}{\cellcolor[HTML]{FFFFFF}{\color[HTML]{000000} \textbf{0.700}}} &
  \multicolumn{1}{c|}{\cellcolor[HTML]{FFFFFF}{\color[HTML]{000000} 0.353}} &
  \multicolumn{1}{c|}{\cellcolor[HTML]{FFFFFF}{\color[HTML]{000000} 0.518}} &
  \multicolumn{1}{c|}{\cellcolor[HTML]{FFFFFF}{\color[HTML]{000000} 0.741}} &
  \multicolumn{1}{c|}{\cellcolor[HTML]{FFFFFF}{\color[HTML]{000000} 0.500}} &
  \multicolumn{1}{c|}{\cellcolor[HTML]{FFFFFF}{\color[HTML]{000000} 0.567}} &
  \multicolumn{1}{c|}{\cellcolor[HTML]{FFFFFF}{\color[HTML]{000000} 0.124}} &
  \multicolumn{1}{c|}{\cellcolor[HTML]{FFFFFF}{\color[HTML]{000000} 0.553   ± 0.21}} \\ \hline
\rowcolor[HTML]{FFFFFF} 
\multicolumn{1}{|l|}{\cellcolor[HTML]{FFFFFF}{\color[HTML]{000000} ARIMA (M)*}} &
  \multicolumn{1}{c|}{\cellcolor[HTML]{FFFFFF}{\color[HTML]{000000} \textbf{0.457}}} &
  \multicolumn{1}{c|}{\cellcolor[HTML]{FFFFFF}{\color[HTML]{000000} \textbf{0.359}}} &
  \multicolumn{1}{c|}{\cellcolor[HTML]{FFFFFF}{\color[HTML]{000000} \textbf{0.752}}} &
  \multicolumn{1}{c|}{\cellcolor[HTML]{FFFFFF}{\color[HTML]{000000} \textbf{0.809}}} &
  \multicolumn{1}{c|}{\cellcolor[HTML]{FFFFFF}{\color[HTML]{000000} 0.807}} &
  \multicolumn{1}{c|}{\cellcolor[HTML]{FFFFFF}{\color[HTML]{000000} 0.690}} &
  \multicolumn{1}{c|}{\cellcolor[HTML]{FFFFFF}{\color[HTML]{000000} \textbf{0.500}}} &
  \multicolumn{1}{c|}{\cellcolor[HTML]{FFFFFF}{\color[HTML]{000000} 0.518}} &
  \multicolumn{1}{c|}{\cellcolor[HTML]{FFFFFF}{\color[HTML]{000000} \textbf{0.769}}} &
  \multicolumn{1}{c|}{\cellcolor[HTML]{FFFFFF}{\color[HTML]{000000} 0.500}} &
  \multicolumn{1}{c|}{\cellcolor[HTML]{FFFFFF}{\color[HTML]{000000} \textbf{0.576}}} &
  \multicolumn{1}{c|}{\cellcolor[HTML]{FFFFFF}{\color[HTML]{000000} \textbf{0.148}}} &
  \multicolumn{1}{c|}{\cellcolor[HTML]{FFFFFF}{\color[HTML]{000000} \textbf{0.574   ± 0.19}}} \\ \hline\hline
\rowcolor[HTML]{FFFFFF} 
\multicolumn{1}{|l|}{\cellcolor[HTML]{FFFFFF}{\color[HTML]{000000} LSTM-DT}} &
  \multicolumn{1}{c|}{\cellcolor[HTML]{FFFFFF}{\color[HTML]{000000} 0.515}} &
  \multicolumn{1}{c|}{\cellcolor[HTML]{FFFFFF}{\color[HTML]{000000} 0.707}} &
  \multicolumn{1}{c|}{\cellcolor[HTML]{FFFFFF}{\color[HTML]{000000} 0.721}} &
  \multicolumn{1}{c|}{\cellcolor[HTML]{FFFFFF}{\color[HTML]{000000} 0.980}} &
  \multicolumn{1}{c|}{\cellcolor[HTML]{FFFFFF}{\color[HTML]{000000} 0.744}} &
  \multicolumn{1}{c|}{\cellcolor[HTML]{FFFFFF}{\color[HTML]{000000} 0.638}} &
  \multicolumn{1}{c|}{\cellcolor[HTML]{FFFFFF}{\color[HTML]{000000} 0.400}} &
  \multicolumn{1}{c|}{\cellcolor[HTML]{FFFFFF}{\color[HTML]{000000} 0.513}} &
  \multicolumn{1}{c|}{\cellcolor[HTML]{FFFFFF}{\color[HTML]{000000} 0.741}} &
  \multicolumn{1}{c|}{\cellcolor[HTML]{FFFFFF}{\color[HTML]{000000} 0.667}} &
  \multicolumn{1}{c|}{\cellcolor[HTML]{FFFFFF}{\color[HTML]{000000} 0.580}} &
  \multicolumn{1}{c|}{\cellcolor[HTML]{FFFFFF}{\color[HTML]{000000} 0.391}} &
  \multicolumn{1}{c|}{\cellcolor[HTML]{FFFFFF}{\color[HTML]{000000} 0.633 ± 0.16}} \\ \hline
\rowcolor[HTML]{FFFFFF} 
\multicolumn{1}{|l|}{\cellcolor[HTML]{FFFFFF}{\color[HTML]{000000} LSTM-DT (M)*}} &
  \multicolumn{1}{c|}{\cellcolor[HTML]{FFFFFF}{\color[HTML]{000000} \textbf{0.521}}} &
  \multicolumn{1}{c|}{\cellcolor[HTML]{FFFFFF}{\color[HTML]{000000} \textbf{0.754}}} &
  \multicolumn{1}{c|}{\cellcolor[HTML]{FFFFFF}{\color[HTML]{000000} 0.729}} &
  \multicolumn{1}{c|}{\cellcolor[HTML]{FFFFFF}{\color[HTML]{000000} \textbf{0.987}}} &
  \multicolumn{1}{c|}{\cellcolor[HTML]{FFFFFF}{\color[HTML]{000000} 0.734}} &
  \multicolumn{1}{c|}{\cellcolor[HTML]{FFFFFF}{\color[HTML]{000000} 0.638}} &
  \multicolumn{1}{c|}{\cellcolor[HTML]{FFFFFF}{\color[HTML]{000000} \textbf{0.600}}} &
  \multicolumn{1}{c|}{\cellcolor[HTML]{FFFFFF}{\color[HTML]{000000} 0.513}} &
  \multicolumn{1}{c|}{\cellcolor[HTML]{FFFFFF}{\color[HTML]{000000} 0.769}} &
  \multicolumn{1}{c|}{\cellcolor[HTML]{FFFFFF}{\color[HTML]{000000} \textbf{0.686}}} &
  \multicolumn{1}{c|}{\cellcolor[HTML]{FFFFFF}{\color[HTML]{000000} \textbf{0.588}}} &
  \multicolumn{1}{c|}{\cellcolor[HTML]{FFFFFF}{\color[HTML]{000000} \textbf{0.446}}} &
  \multicolumn{1}{c|}{\cellcolor[HTML]{FFFFFF}{\color[HTML]{000000} 0.664   ± 0.14}} \\ \hline
\rowcolor[HTML]{FFFFFF} 
\multicolumn{1}{|l|}{\cellcolor[HTML]{FFFFFF}{\color[HTML]{000000} LSTM-DT (M, Bi)*}} &
  \multicolumn{1}{c|}{\cellcolor[HTML]{FFFFFF}{\color[HTML]{000000} 0.505}} &
  \multicolumn{1}{c|}{\cellcolor[HTML]{FFFFFF}{\color[HTML]{000000} 0.662}} &
  \multicolumn{1}{c|}{\cellcolor[HTML]{FFFFFF}{\color[HTML]{000000} \textbf{0.755}}} &
  \multicolumn{1}{c|}{\cellcolor[HTML]{FFFFFF}{\color[HTML]{000000} 0.949}} &
  \multicolumn{1}{c|}{\cellcolor[HTML]{FFFFFF}{\color[HTML]{000000} \textbf{0.895}}} &
  \multicolumn{1}{c|}{\cellcolor[HTML]{FFFFFF}{\color[HTML]{000000} \textbf{0.792}}} &
  \multicolumn{1}{c|}{\cellcolor[HTML]{FFFFFF}{\color[HTML]{000000} 0.545}} &
  \multicolumn{1}{c|}{\cellcolor[HTML]{FFFFFF}{\color[HTML]{000000} 0.488}} &
  \multicolumn{1}{c|}{\cellcolor[HTML]{FFFFFF}{\color[HTML]{000000} \textbf{0.786}}} &
  \multicolumn{1}{c|}{\cellcolor[HTML]{FFFFFF}{\color[HTML]{000000} 0.684}} &
  \multicolumn{1}{c|}{\cellcolor[HTML]{FFFFFF}{\color[HTML]{000000} 0.587}} &
  \multicolumn{1}{c|}{\cellcolor[HTML]{FFFFFF}{\color[HTML]{000000} 0.432}} &
  \multicolumn{1}{c|}{\cellcolor[HTML]{FFFFFF}{\color[HTML]{000000} \textbf{0.673   ± 0.16}}} \\ \hline\hline
\rowcolor[HTML]{FFFFFF} 
\multicolumn{1}{|l|}{\cellcolor[HTML]{FFFFFF}{\color[HTML]{000000} LSTM-AE}} &
  \multicolumn{1}{c|}{\cellcolor[HTML]{FFFFFF}{\color[HTML]{000000} 0.500}} &
  \multicolumn{1}{c|}{\cellcolor[HTML]{FFFFFF}{\color[HTML]{000000} \textbf{0.705}}} &
  \multicolumn{1}{c|}{\cellcolor[HTML]{FFFFFF}{\color[HTML]{000000} 0.610}} &
  \multicolumn{1}{c|}{\cellcolor[HTML]{FFFFFF}{\color[HTML]{000000} 0.866}} &
  \multicolumn{1}{c|}{\cellcolor[HTML]{FFFFFF}{\color[HTML]{000000} 0.420}} &
  \multicolumn{1}{c|}{\cellcolor[HTML]{FFFFFF}{\color[HTML]{000000} \textbf{0.253}}} &
  \multicolumn{1}{c|}{\cellcolor[HTML]{FFFFFF}{\color[HTML]{000000} 0.545}} &
  \multicolumn{1}{c|}{\cellcolor[HTML]{FFFFFF}{\color[HTML]{000000} 0.750}} &
  \multicolumn{1}{c|}{\cellcolor[HTML]{FFFFFF}{\color[HTML]{000000} \textbf{0.692}}} &
  \multicolumn{1}{c|}{\cellcolor[HTML]{FFFFFF}{\color[HTML]{000000} 0.457}} &
  \multicolumn{1}{c|}{\cellcolor[HTML]{FFFFFF}{\color[HTML]{000000} 0.483}} &
  \multicolumn{1}{c|}{\cellcolor[HTML]{FFFFFF}{\color[HTML]{000000} 0.314}} &
  \multicolumn{1}{c|}{\cellcolor[HTML]{FFFFFF}{\color[HTML]{000000} 0.550 ± 0.17}} \\ \hline
\rowcolor[HTML]{FFFFFF} 
\multicolumn{1}{|l|}{\cellcolor[HTML]{FFFFFF}{\color[HTML]{000000} LSTM-AE (M)*}} &
  \multicolumn{1}{c|}{\cellcolor[HTML]{FFFFFF}{\color[HTML]{000000} \textbf{0.522}}} &
  \multicolumn{1}{c|}{\cellcolor[HTML]{FFFFFF}{\color[HTML]{000000} 0.701}} &
  \multicolumn{1}{c|}{\cellcolor[HTML]{FFFFFF}{\color[HTML]{000000} \textbf{0.644}}} &
  \multicolumn{1}{c|}{\cellcolor[HTML]{FFFFFF}{\color[HTML]{000000} \textbf{0.882}}} &
  \multicolumn{1}{c|}{\cellcolor[HTML]{FFFFFF}{\color[HTML]{000000} \textbf{0.442}}} &
  \multicolumn{1}{c|}{\cellcolor[HTML]{FFFFFF}{\color[HTML]{000000} 0.236}} &
  \multicolumn{1}{c|}{\cellcolor[HTML]{FFFFFF}{\color[HTML]{000000} \textbf{0.667}}} &
  \multicolumn{1}{c|}{\cellcolor[HTML]{FFFFFF}{\color[HTML]{000000} 0.750}} &
  \multicolumn{1}{c|}{\cellcolor[HTML]{FFFFFF}{\color[HTML]{000000} 0.609}} &
  \multicolumn{1}{c|}{\cellcolor[HTML]{FFFFFF}{\color[HTML]{000000} \textbf{0.533}}} &
  \multicolumn{1}{c|}{\cellcolor[HTML]{FFFFFF}{\color[HTML]{000000} \textbf{0.542}}} &
  \multicolumn{1}{c|}{\cellcolor[HTML]{FFFFFF}{\color[HTML]{000000} \textbf{0.334}}} &
  \multicolumn{1}{c|}{\cellcolor[HTML]{FFFFFF}{\color[HTML]{000000} \textbf{0.572   ± 0.17}}} \\ \hline\hline
\rowcolor[HTML]{FFFFFF} 
\multicolumn{1}{|l|}{\cellcolor[HTML]{FFFFFF}{\color[HTML]{000000} LSTM-VAE}} &
  \multicolumn{1}{c|}{\cellcolor[HTML]{FFFFFF}{\color[HTML]{000000} \textbf{0.526}}} &
  \multicolumn{1}{c|}{\cellcolor[HTML]{FFFFFF}{\color[HTML]{000000} 0.653}} &
  \multicolumn{1}{c|}{\cellcolor[HTML]{FFFFFF}{\color[HTML]{000000} 0.575}} &
  \multicolumn{1}{c|}{\cellcolor[HTML]{FFFFFF}{\color[HTML]{000000} 0.823}} &
  \multicolumn{1}{c|}{\cellcolor[HTML]{FFFFFF}{\color[HTML]{000000} 0.432}} &
  \multicolumn{1}{c|}{\cellcolor[HTML]{FFFFFF}{\color[HTML]{000000} 0.240}} &
  \multicolumn{1}{c|}{\cellcolor[HTML]{FFFFFF}{\color[HTML]{000000} \textbf{0.667}}} &
  \multicolumn{1}{c|}{\cellcolor[HTML]{FFFFFF}{\color[HTML]{000000} 0.700}} &
  \multicolumn{1}{c|}{\cellcolor[HTML]{FFFFFF}{\color[HTML]{000000} \textbf{0.643}}} &
  \multicolumn{1}{c|}{\cellcolor[HTML]{FFFFFF}{\color[HTML]{000000} 0.483}} &
  \multicolumn{1}{c|}{\cellcolor[HTML]{FFFFFF}{\color[HTML]{000000} 0.590}} &
  \multicolumn{1}{c|}{\cellcolor[HTML]{FFFFFF}{\color[HTML]{000000} 0.317}} &
  \multicolumn{1}{c|}{\cellcolor[HTML]{FFFFFF}{\color[HTML]{000000} 0.554 ± 0.16}} \\ \hline
\rowcolor[HTML]{FFFFFF} 
\multicolumn{1}{|l|}{\cellcolor[HTML]{FFFFFF}{\color[HTML]{000000} LSTM-VAE (M)*}} &
  \multicolumn{1}{c|}{\cellcolor[HTML]{FFFFFF}{\color[HTML]{000000} 0.521}} &
  \multicolumn{1}{c|}{\cellcolor[HTML]{FFFFFF}{\color[HTML]{000000} \textbf{0.710}}} &
  \multicolumn{1}{c|}{\cellcolor[HTML]{FFFFFF}{\color[HTML]{000000} \textbf{0.628}}} &
  \multicolumn{1}{c|}{\cellcolor[HTML]{FFFFFF}{\color[HTML]{000000} \textbf{0.901}}} &
  \multicolumn{1}{c|}{\cellcolor[HTML]{FFFFFF}{\color[HTML]{000000} \textbf{0.460}}} &
  \multicolumn{1}{c|}{\cellcolor[HTML]{FFFFFF}{\color[HTML]{000000} \textbf{0.246}}} &
  \multicolumn{1}{c|}{\cellcolor[HTML]{FFFFFF}{\color[HTML]{000000} 0.545}} &
  \multicolumn{1}{c|}{\cellcolor[HTML]{FFFFFF}{\color[HTML]{000000} \textbf{0.764}}} &
  \multicolumn{1}{c|}{\cellcolor[HTML]{FFFFFF}{\color[HTML]{000000} 0.615}} &
  \multicolumn{1}{c|}{\cellcolor[HTML]{FFFFFF}{\color[HTML]{000000} \textbf{0.519}}} &
  \multicolumn{1}{c|}{\cellcolor[HTML]{FFFFFF}{\color[HTML]{000000} 0.590}} &
  \multicolumn{1}{c|}{\cellcolor[HTML]{FFFFFF}{\color[HTML]{000000} \textbf{0.333}}} &
  \multicolumn{1}{c|}{\cellcolor[HTML]{FFFFFF}{\color[HTML]{000000} \textbf{0.569   ± 0.17}}} \\ \hline\hline
\rowcolor[HTML]{FFFFFF} 
\multicolumn{1}{|l|}{\cellcolor[HTML]{FFFFFF}{\color[HTML]{000000} TadGAN}} &
  \multicolumn{1}{c|}{\cellcolor[HTML]{FFFFFF}{\color[HTML]{000000} 0.584}} &
  \multicolumn{1}{c|}{\cellcolor[HTML]{FFFFFF}{\color[HTML]{000000} 0.617}} &
  \multicolumn{1}{c|}{\cellcolor[HTML]{FFFFFF}{\color[HTML]{000000} 0.533}} &
  \multicolumn{1}{c|}{\cellcolor[HTML]{FFFFFF}{\color[HTML]{000000} 0.842}} &
  \multicolumn{1}{c|}{\cellcolor[HTML]{FFFFFF}{\color[HTML]{000000} 0.391}} &
  \multicolumn{1}{c|}{\cellcolor[HTML]{FFFFFF}{\color[HTML]{000000} \textbf{0.297}}} &
  \multicolumn{1}{c|}{\cellcolor[HTML]{FFFFFF}{\color[HTML]{000000} 0.571}} &
  \multicolumn{1}{c|}{\cellcolor[HTML]{FFFFFF}{\color[HTML]{000000} 0.677}} &
  \multicolumn{1}{c|}{\cellcolor[HTML]{FFFFFF}{\color[HTML]{000000} 0.720}} &
  \multicolumn{1}{c|}{\cellcolor[HTML]{FFFFFF}{\color[HTML]{000000} 0.581}} &
  \multicolumn{1}{c|}{\cellcolor[HTML]{FFFFFF}{\color[HTML]{000000} 0.588}} &
  \multicolumn{1}{c|}{\cellcolor[HTML]{FFFFFF}{\color[HTML]{000000} 0.162}} &
  \multicolumn{1}{c|}{\cellcolor[HTML]{FFFFFF}{\color[HTML]{000000} 0.547 ± 0.18}} \\ \hline
\rowcolor[HTML]{FFFFFF} 
\multicolumn{1}{|l|}{\cellcolor[HTML]{FFFFFF}{\color[HTML]{000000} TadGAN (M)*}} &
  \multicolumn{1}{c|}{\cellcolor[HTML]{FFFFFF}{\color[HTML]{000000} 0.584}} &
  \multicolumn{1}{c|}{\cellcolor[HTML]{FFFFFF}{\color[HTML]{000000} \textbf{0.630}}} &
  \multicolumn{1}{c|}{\cellcolor[HTML]{FFFFFF}{\color[HTML]{000000} \textbf{0.534}}} &
  \multicolumn{1}{c|}{\cellcolor[HTML]{FFFFFF}{\color[HTML]{000000} \textbf{0.846}}} &
  \multicolumn{1}{c|}{\cellcolor[HTML]{FFFFFF}{\color[HTML]{000000} \textbf{0.395}}} &
  \multicolumn{1}{c|}{\cellcolor[HTML]{FFFFFF}{\color[HTML]{000000} 0.291}} &
  \multicolumn{1}{c|}{\cellcolor[HTML]{FFFFFF}{\color[HTML]{000000} \textbf{0.615}}} &
  \multicolumn{1}{c|}{\cellcolor[HTML]{FFFFFF}{\color[HTML]{000000} 0.677}} &
  \multicolumn{1}{c|}{\cellcolor[HTML]{FFFFFF}{\color[HTML]{000000} 0.720}} &
  \multicolumn{1}{c|}{\cellcolor[HTML]{FFFFFF}{\color[HTML]{000000} 0.581}} &
  \multicolumn{1}{c|}{\cellcolor[HTML]{FFFFFF}{\color[HTML]{000000} 0.588}} &
  \multicolumn{1}{c|}{\cellcolor[HTML]{FFFFFF}{\color[HTML]{000000} \textbf{0.164}}} &
  \multicolumn{1}{c|}{\cellcolor[HTML]{FFFFFF}{\color[HTML]{000000} \textbf{0.552   ± 0.18}}} \\ \hline
{\color[HTML]{000000} } &
  \multicolumn{1}{l}{{\color[HTML]{000000} }} &
  \multicolumn{1}{l}{{\color[HTML]{000000} }} &
  \multicolumn{1}{l}{{\color[HTML]{000000} }} &
  \multicolumn{1}{l}{{\color[HTML]{000000} }} &
  \multicolumn{1}{l}{{\color[HTML]{000000} }} &
  \multicolumn{1}{l}{{\color[HTML]{000000} }} &
  \multicolumn{1}{l}{{\color[HTML]{000000} }} &
  \multicolumn{1}{l}{{\color[HTML]{000000} }} &
  \multicolumn{1}{l}{{\color[HTML]{000000} }} &
  \multicolumn{1}{l}{{\color[HTML]{000000} }} &
  \multicolumn{1}{l}{{\color[HTML]{000000} }} &
  \multicolumn{1}{l}{{\color[HTML]{000000} }} &
  \multicolumn{1}{l}{{\color[HTML]{000000} }} \\
\multicolumn{14}{c}{{\color[HTML]{000000} \textbf{B: Ablation Study}}} \\ \hline
\rowcolor[HTML]{E7E6E6} 
\multicolumn{1}{|c|}{\cellcolor[HTML]{E7E6E6}{\color[HTML]{000000} }} &
  \multicolumn{2}{c|}{\cellcolor[HTML]{E7E6E6}{\color[HTML]{000000} \textbf{NASA}}} &
  \multicolumn{4}{c|}{\cellcolor[HTML]{E7E6E6}{\color[HTML]{000000} \textbf{YAHOO}}} &
  \multicolumn{5}{c|}{\cellcolor[HTML]{E7E6E6}{\color[HTML]{000000} \textbf{NAB}}} &
  \multicolumn{1}{c|}{\cellcolor[HTML]{E7E6E6}{\color[HTML]{000000} \textbf{UCR}}} &
  \multicolumn{1}{c|}{\cellcolor[HTML]{E7E6E6}{\color[HTML]{000000} }} \\ \cline{2-13}
\rowcolor[HTML]{FFF2CC} 
\multicolumn{1}{|c|}{\multirow{-2}{*}{\cellcolor[HTML]{E7E6E6}{\color[HTML]{000000} \textbf{Models}}}} &
  \multicolumn{1}{c|}{\cellcolor[HTML]{FFF2CC}{\color[HTML]{000000} \textbf{MSL}}} &
  \multicolumn{1}{c|}{\cellcolor[HTML]{FFF2CC}{\color[HTML]{000000} \textbf{SMAP}}} &
  \multicolumn{1}{c|}{\cellcolor[HTML]{FFF2CC}{\color[HTML]{000000} \textbf{A1}}} &
  \multicolumn{1}{c|}{\cellcolor[HTML]{FFF2CC}{\color[HTML]{000000} \textbf{A2}}} &
  \multicolumn{1}{c|}{\cellcolor[HTML]{FFF2CC}{\color[HTML]{000000} \textbf{A3}}} &
  \multicolumn{1}{c|}{\cellcolor[HTML]{FFF2CC}{\color[HTML]{000000} \textbf{A4}}} &
  \multicolumn{1}{c|}{\cellcolor[HTML]{FFF2CC}{\color[HTML]{000000} \textbf{Art}}} &
  \multicolumn{1}{c|}{\cellcolor[HTML]{FFF2CC}{\color[HTML]{000000} \textbf{AdEx}}} &
  \multicolumn{1}{c|}{\cellcolor[HTML]{FFF2CC}{\color[HTML]{000000} \textbf{AWS}}} &
  \multicolumn{1}{c|}{\cellcolor[HTML]{FFF2CC}{\color[HTML]{000000} \textbf{Traffic}}} &
  \multicolumn{1}{c|}{\cellcolor[HTML]{FFF2CC}{\color[HTML]{000000} \textbf{Tweets}}} &
  \multicolumn{1}{c|}{\cellcolor[HTML]{FFF2CC}{\color[HTML]{000000} \textbf{UCR}}} &
  \multicolumn{1}{c|}{\multirow{-2}{*}{\cellcolor[HTML]{E7E6E6}{\color[HTML]{000000} \textbf{Avg. F1 ($\mu \pm \sigma$)}}}} \\ \hline
\multicolumn{1}{|l|}{\cellcolor[HTML]{FFFFFF}{\color[HTML]{000000} AER (PRED)*}} &
  \multicolumn{1}{c|}{{\color[HTML]{000000} 0.494}} &
  \multicolumn{1}{c|}{{\color[HTML]{000000} 0.685}} &
  \multicolumn{1}{c|}{{\color[HTML]{000000} 0.705}} &
  \multicolumn{1}{c|}{\cellcolor[HTML]{FFFFFF}{\color[HTML]{000000} 0.923}} &
  \multicolumn{1}{c|}{\cellcolor[HTML]{E2EFDA}{\color[HTML]{000000} \textbf{0.896}}} &
  \multicolumn{1}{c|}{\cellcolor[HTML]{E2EFDA}{\color[HTML]{000000} \textbf{0.722}}} &
  \multicolumn{1}{c|}{\cellcolor[HTML]{FFFFFF}{\color[HTML]{000000} 0.500}} &
  \multicolumn{1}{c|}{\cellcolor[HTML]{FFFFFF}{\color[HTML]{000000} 0.541}} &
  \multicolumn{1}{c|}{\cellcolor[HTML]{FFFFFF}{\color[HTML]{000000} 0.688}} &
  \multicolumn{1}{c|}{\cellcolor[HTML]{FFFFFF}{\color[HTML]{000000} \textbf{0.615}}} &
  \multicolumn{1}{c|}{\cellcolor[HTML]{FFFFFF}{\color[HTML]{000000} 0.556}} &
  \multicolumn{1}{c|}{{\color[HTML]{000000} 0.461}} &
  \multicolumn{1}{c|}{\cellcolor[HTML]{FFFFFF}{\color[HTML]{000000} 0.649   ± 0.14}} \\ \hline
\multicolumn{1}{|l|}{\cellcolor[HTML]{FFFFFF}{\color[HTML]{000000} AER (SUM)*}} &
  \multicolumn{1}{c|}{{\color[HTML]{000000} 0.488}} &
  \multicolumn{1}{c|}{{\color[HTML]{000000} 0.680}} &
  \multicolumn{1}{c|}{{\color[HTML]{000000} 0.714}} &
  \multicolumn{1}{c|}{\cellcolor[HTML]{FFFFFF}{\color[HTML]{000000} 0.936}} &
  \multicolumn{1}{c|}{{\color[HTML]{000000} 0.719}} &
  \multicolumn{1}{c|}{{\color[HTML]{000000} 0.553}} &
  \multicolumn{1}{c|}{{\color[HTML]{000000} 0.500}} &
  \multicolumn{1}{c|}{\cellcolor[HTML]{FFFFFF}{\color[HTML]{000000} \textbf{0.702}}} &
  \multicolumn{1}{c|}{\cellcolor[HTML]{FFFFFF}{\color[HTML]{000000} \textbf{0.733}}} &
  \multicolumn{1}{c|}{\cellcolor[HTML]{FFFFFF}{\color[HTML]{000000} 0.606}} &
  \multicolumn{1}{c|}{{\color[HTML]{000000} 0.559}} &
  \multicolumn{1}{c|}{{\color[HTML]{000000} 0.428}} &
  \multicolumn{1}{c|}{\cellcolor[HTML]{FFFFFF}{\color[HTML]{000000} 0.635   ± 0.13}} \\ \hline
\multicolumn{1}{|l|}{\cellcolor[HTML]{FFFFFF}{\color[HTML]{000000} AER (REC)*}} &
  \multicolumn{1}{c|}{{\color[HTML]{000000} 0.500}} &
  \multicolumn{1}{c|}{{\color[HTML]{000000} 0.683}} &
  \multicolumn{1}{c|}{{\color[HTML]{000000} 0.707}} &
  \multicolumn{1}{c|}{\cellcolor[HTML]{FFFFFF}{\color[HTML]{000000} \textbf{0.985}}} &
  \multicolumn{1}{c|}{{\color[HTML]{000000} 0.620}} &
  \multicolumn{1}{c|}{{\color[HTML]{000000} 0.416}} &
  \multicolumn{1}{c|}{{\color[HTML]{000000} 0.444}} &
  \multicolumn{1}{c|}{{\color[HTML]{000000} 0.644}} &
  \multicolumn{1}{c|}{{\color[HTML]{000000} 0.692}} &
  \multicolumn{1}{c|}{{\color[HTML]{000000} 0.571}} &
  \multicolumn{1}{c|}{{\color[HTML]{000000} 0.519}} &
  \multicolumn{1}{c|}{{\color[HTML]{000000} 0.363}} &
  \multicolumn{1}{c|}{\cellcolor[HTML]{FFFFFF}{\color[HTML]{000000} 0.595   ± 0.16}} \\ \hline
\multicolumn{1}{|l|}{\cellcolor[HTML]{FFFFFF}{\color[HTML]{000000} AER (MULT)*}} &
  \multicolumn{1}{c|}{\cellcolor[HTML]{E2EFDA}{\color[HTML]{000000} \textbf{0.541}}} &
  \multicolumn{1}{c|}{\cellcolor[HTML]{E2EFDA}{\color[HTML]{000000} \textbf{0.772}}} &
  \multicolumn{1}{c|}{\cellcolor[HTML]{E2EFDA}{\color[HTML]{000000} \textbf{0.772}}} &
  \multicolumn{1}{c|}{\cellcolor[HTML]{E2EFDA}{\color[HTML]{000000} 0.959}} &
  \multicolumn{1}{c|}{{\color[HTML]{000000} 0.752}} &
  \multicolumn{1}{c|}{{\color[HTML]{000000} 0.572}} &
  \multicolumn{1}{c|}{\cellcolor[HTML]{E2EFDA}{\color[HTML]{000000} \textbf{0.615}}} &
  \multicolumn{1}{c|}{\cellcolor[HTML]{E2EFDA}{\color[HTML]{000000} 0.635}} &
  \multicolumn{1}{c|}{\cellcolor[HTML]{E2EFDA}{\color[HTML]{000000} 0.621}} &
  \multicolumn{1}{c|}{\cellcolor[HTML]{E2EFDA}{\color[HTML]{000000} 0.606}} &
  \multicolumn{1}{c|}{\cellcolor[HTML]{E2EFDA}{\color[HTML]{000000} \textbf{0.585}}} &
  \multicolumn{1}{c|}{\cellcolor[HTML]{E2EFDA}{\color[HTML]{000000} \textbf{0.470}}} &
  \multicolumn{1}{c|}{\cellcolor[HTML]{FFFFFF}{\color[HTML]{000000} \textbf{0.658   ± 0.13}}} \\ \hline
\end{tabular}
}
\vspace{.20cm}
\caption{F1 scores for masking, bi-directional scoring, and ablation study. All new methods are marked with *. \bl{The variation with the highest score is in \textbf{bold} for each model type. The final AER model uses the green fill scores.}}
\label{5:other_scores}
\end{table*}

\subsection{Benchmarking Results}

\textbf{AER outperforms baseline models based on averaged F1 scores (RQ1).} Table \ref{5:all_scores} shows that AER has an averaged F1 score of \bl{0.683}, which is \bl{23.5\%} higher than the score of the standard ARIMA model. The flexibility of combining prediction-based and reconstruction-based anomaly scores leads to an improvement in F1 scores across the datasets. The graph in Fig.~\ref{5:runtime} shows the runtime of AER \bl{scales} in the same order as LSTM-DT, LSTM-AE, and LSTM-VAE. While the runtime is slighter higher for AER than for those models, this is a very reasonable computation cost considering the performance increase.

\textbf{AER v.s. TadGAN (RQ1)}. Similarly, Table \ref{5:all_scores} shows that AER outperforms TadGAN by \bl{24.9\%} in terms of averaged F1 scores while requiring \bl{less execution time} (see Fig.~\ref{5:runtime}). This result suggests that combining prediction-based with reconstruction-based anomaly scores could lead to better F1 scores than combining critic-based with reconstruction-based anomaly scores.

\textbf{Masking improves averaged F1 scores slightly (RQ2)}. Table \ref{5:other_scores}-A shows that masking scores improved averaged F1 scores by 4.3\%, on average, for prediction-based methods and 2.6\%, on average, for reconstruction-based methods. Masking anomaly scores benefited prediction-based methods more than reconstruction-based methods since those methods tend to make more false-positive predictions. However, masking may remove anomalies at the start of the signal and hurt model performance on datasets like \texttt{YAHOOA3} and \texttt{YAHOOA4}.

\textbf{Bi-directional scoring greatly improves F1 scores on some datasets (RQ2)}. \bl{LSTM-DT (M, Bi) consists of two separate LSTM-DT models trained on the sequence in the forward and reversed direction respectively.} Table \ref{5:all_scores}-A shows that using bi-directional scoring with LSTM-DT (M, Bi) improved F1 scores by \bl{20.3}\% for the \texttt{YAHOOA3} dataset and 24.1\% for the \texttt{YAHOOA4} dataset compared to LSTM-DT. These datasets have signals with point anomalies at the beginning that uni-directional prediction-based models cannot predict. However, bi-directional scoring may negatively impact the performance of models on other datasets. Since prediction-based methods tend to produce false-positive predictions, filling in anomaly scores missed by prediction-based anomaly scores allows for more opportunities to produce false positives.

\subsection{Ablation Study}



\bl{\textbf{Product (MULT) combination of anomaly scores have the highest averaged F1 score across all combination methods (RQ3)}. The product (MULT) combination of prediction-based and reconstruction-based anomaly scores produced the highest F1 scores on 6 of 12 datasets (see Table~\ref{5:other_scores}-B). Most of these datasets were non-synthetic, including \texttt{MSL}, \texttt{SMAP}, \texttt{YAHOOA1}, and \texttt{Tweets}. This combination method outperformed the convex (SUM) combination by 3.7\%, the reconstruction-based only (REC) combination by 10.6\%, and the prediction-based only (PRED) combination by 1.5\% in terms of averaged F1 scores. Additionally, excluding \texttt{YAHOOA3} and \texttt{YAHOOA4} synthetic datasets with many point anomalies result in an averaged F1 score of 0.658 for the product (MULT) combination, a 6.6\% increase compared to 0.617 for prediction-based only (PRED) combination. These results support the idea that mixture anomaly scores offer more information than reconstruction-based anomaly scores in general and prediction-based anomaly scores in cases other than identifying point anomalies.}

\bl{\textbf{Prediction-based only (PRED) anomaly scores perform better on datasets with mostly point anomalies.} Bi-directional scoring produced the highest F1 scores on datasets like \texttt{YAHOOA3} and \texttt{YAHOOA4} with mostly point anomalies (see Table~\ref{5:other_scores}-B). This finding is consistent with our findings in the LSTM-DT (M, Bi) model.}

\bl{\textbf{The selection of the combination method for each dataset is based on the use case}. We recommend that users default to using product (MULT) anomaly scores and using prediction-based only (PRED) scores when users primarily want to identify point anomalies. The AER model reports the F1 scores of AER (PRED) for the \texttt{YAHOOA3} and \texttt{YAHOOA4} datasets with mostly point anomalies and AER (MULT) for the other datasets, even though they might not be the best combination method according to the ablation study. In practice, datasets come without labels since anomaly detection is an unsupervised problem. Hence, it is impossible to retroactively tune the best method to calculate anomaly scores for each dataset.}

\subsection{Limitations and Discussion}

While product mixture scores offer unique insights for anomaly detection, several ways exist to improve the AER framework. For example, the model architecture could be better since our study uses a vanilla auto-encoder architecture with one biLSTM layer for both the encoder and decoder. \bl{Our} framework is designed to easily extend to any reconstruction-based method with minimum changes to the objective function. Another improvement involves \bl{experimenting with} the $\gamma$ (defaults to 0.5), which controls the contribution of prediction and reconstruction loss to the objective function. An optimal $\gamma$ could lead to more accurate prediction-based and reconstruction-based anomaly scores that ultimately improve F1 scores. Lastly, the findings in our analysis of existing methods in section IV are for datasets we are currently investigating. The identified constraints may not always hold in other datasets. 

Although researchers pay increasing attention to building more powerful models to improve the accuracy of prediction-based and reconstruction-based methods, we would like to call for more attention to the post-processing stage. Our study demonstrated that changes in the post-processing stage could significantly improve performances in addition to our proposed model. Future exploration directions could include additional methods to create mixture scores and \bl{better heuristics for the} selection of such methods (e.g., between PRED and MULT) for each signal.


%% file: 6_conclusion.tex
\section{Conclusion}

This study analyzed the successes and limitations of existing reconstruction-based and prediction-based methods. We proposed a threefold solution to address existing limitations: (1) the AER framework that leverages the successes of prediction-based and reconstruction-based methods, (2) masking anomaly scores to reduce start-of-sequence false-positive predictions, and (3) bi-directional scoring to address missing forecast issues. In addition, we conducted an ablation study to test several ways of combining prediction-based and reconstruction-based anomaly scores. Our results showed that (1) AER has the highest \bl{F1 score averaged across 12 datasets}, (2) masking and bi-directional scoring improve F1 scores given the right conditions, (3) the product combination (MULT) of bi-directional and reconstruction-based anomaly scores produces better results, on average, for \bl{datasets with mostly collective anomalies}. Finally, the code is available at \url{https://github.com/sintel-dev/Orion}.

